\documentclass[lettersize,journal]{IEEEtran}
\usepackage{amsmath,amsfonts}
\usepackage{algorithmic}
\usepackage{array}
\usepackage[font=normalsize,labelfont=sf,textfont=sf]{subcaption}
\usepackage{textcomp}
\usepackage{stfloats}
\usepackage{url}
\usepackage{verbatim}
\usepackage{graphicx}
\def\BibTeX{{\rm B\kern-.05em{\sc i\kern-.025em b}\kern-.08em
    T\kern-.1667em\lower.7ex\hbox{E}\kern-.125emX}}
\usepackage{balance}

\usepackage{nicefrac}
\usepackage{booktabs}
\usepackage{multirow}
\usepackage[table]{xcolor}

\usepackage{gensymb}
\usepackage{microtype}
\usepackage{enumitem}
\usepackage{cite}
\definecolor{citecolor}{RGB}{119,185,0} 
\usepackage[pagebackref=false,breaklinks=true,colorlinks,citecolor=citecolor,bookmarks=false]{hyperref}

\newcommand{\best}[1]{\textbf{#1}}
\newcommand{\second}[1]{\underline{#1}}
\newcommand{\gain}[1]{\textcolor{red}{\,(#1)}}
\newcommand{\same}[1]{\textcolor{black}{\scriptsize{\,(#1)}}}

\begin{document}
\title{UniGeo: A Multi-modal Large Language Model for Text-Guided Cross-View Geo-Localization}
\author{Jiahao Wen, Hang Yu, ~\IEEEmembership{Member, ~IEEE}, and Zhedong Zheng, ~\IEEEmembership{Senior Member, ~IEEE} 
\thanks{Manuscript received xxx; accepted . date of publication xxx; date of current version xxx. (Corresponding author: Hang Yu.)}
\thanks{Jiahao Wen and Hang Yu are with the School of Computer Engineering and Science, Shanghai University, 
Shanghai 200444, China (e-mail: wenjh@shu.edu.cn; yuhang@shu.edu.cn). }
\thanks{Zhedong Zheng is with the Faculty of Science and Technology, and Institute of Collaborative Innovation, University of Macau, Macau 999078, China (e-mail: zhedongzheng@um.edu.mo).}}

\markboth{Journal of \LaTeX\ Class Files,~Vol.~18, No.~9, September~2020}%
{How to Use the IEEEtran \LaTeX \ Templates}

\maketitle
\begin{abstract}
Text-guided drone geo-localization aims to localize a target region in a large-scale image gallery based on a natural language description. Current methods mostly formulate this task as direct matching between an open-ended language query and candidate images. However, when the query is incomplete and candidate images are highly similar, relying solely on global cross-modal matching is often insufficient for reliable fine-grained localization. 
Recent progress in  multi-modal large language models (MLLMs) offers a promising foundation for addressing these challenges. 
In this paper, we propose UniGeo, a unified MLLM for text-guided drone geo-localization. UniGeo is built upon a shared vision-language modeling framework and jointly supports geo-semantic understanding, cross-view semantic generation, and candidate-level fine-grained verification.
Specifically, we first establish stable correspondences among local scene elements, spatial structural relations, and language descriptions through geo-semantic understanding learning.
We then further model the semantic mapping between drone and satellite views through cross-view generation learning. Building on this foundation, we introduce a plug-and-play verification module to perform fine-grained discrimination over highly confusable candidates.
In addition, we design a multi-stage training framework that decouples geo-semantic learning, cross-view discriminative enhancement, and candidate-level verification modeling, thereby improving the adaptability of the unified model to text-guided drone geo-localization and strengthening its fine-grained geo-localization capability. 
Experimental results show that UniGeo achieves stable performance improvements across multiple retrieval backbones. In particular, on the text-guided geo-localization task, UniGeo improves GeoText-1652 by $+ 13.59\%$ and $+ 2.83\%$ in R@10 and mAP, validating the effectiveness of unified vision-language modeling for fine-grained text-guided drone geo-localization.
\end{abstract}




\section{Introduction}
Drone geo-localization addresses the problem of identifying the geographic location corresponding to a target region from a candidate gallery, and plays a critical role in linking local scene cues with large-scale geospatial context~\cite{Zheng_Wei_Yang_202001,Wang_Zheng_Yan_Zhang_Sun_Zheng_Yang_202202, Lin_Zheng_Zhong_Luo_Li_Yang_Sebe_202203, 968495004,wen2025WeatherPrompt05}. Existing research has largely focused on image-query settings, where a drone reference image is matched against a cross-platform candidate gallery to retrieve the target region~\cite{wang2024Muse07,deuser2023sample4geo08,lin2024self6209,du2024ccr6310}. In practical applications, however, standardized reference images are often unavailable or difficult to acquire reliably. By contrast, users are more likely to specify the target region through natural language, describing its appearance, spatial layout, and relative relations to surrounding structures~\cite{chai2026like}. This makes text-guided drone geo-localization a practically important problem with clear real-world relevance.

\begin{figure}[t]
    \centering
    \includegraphics[width=0.95\linewidth]{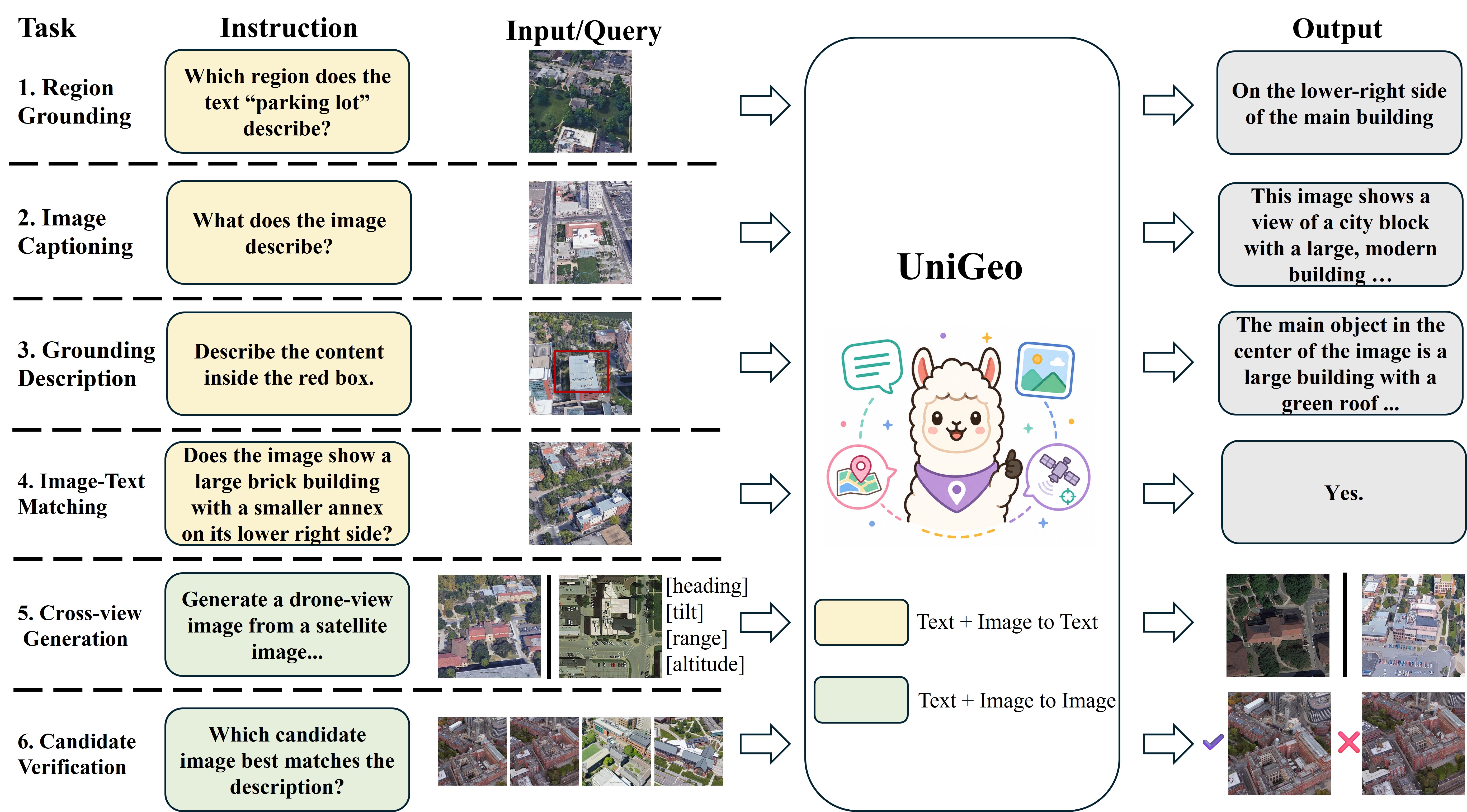}
    \caption{\textbf{Unified multi-task capability of UniGeo.}
UniGeo is a unified MLLM for text-guided geo-localization, supporting multiple geo-localization-oriented tasks such as region grounding, image captioning, region-grounded description, image-text matching, pose-aware cross-view generation, and candidate verification.}
    \label{fig1}
    \vspace{-.20in}
\end{figure}

Recent studies have begun to explore text-guided drone geo-localization~\cite{GeoText165206,ye2025cross11,ji2025mmgeo12,yuan2025seeing13}. However, most existing methods still formulate it as direct matching based on the original natural language query, implicitly assuming that the query itself already provides sufficiently stable and discriminative evidence for geo-localization, an assumption that is often violated in practice~\cite{hu2023cross}.
This task presents two primary challenges. 
\textbf{(1) Text Query Underspecification.} Open-ended natural language descriptions are inherently selective and incomplete, typically capturing only a subset of the salient cues of the target region, and are therefore often insufficient to provide the fine-grained structural information and stable semantic constraints required for distinguishing among highly similar candidates. Moreover, the same target region can be described at different levels of granularity and with different semantic emphases, which further increases the uncertainty of semantic modeling.
\textbf{(2) Candidate Disambiguation Difficulty.} Even when the query explicitly specifies spatial layouts, relative relations, or surrounding cues, the intended meaning of such information often cannot be resolved accurately without reference to specific candidate scenes. This issue is particularly pronounced in drone scenarios, where language descriptions are typically derived from local, selective, and task-driven observations, whereas candidate images usually come from different platforms and viewpoints with substantially different spatial organizations and semantic manifestations.

\begin{figure}[t]
    \centering
    \includegraphics[width=0.9\linewidth]{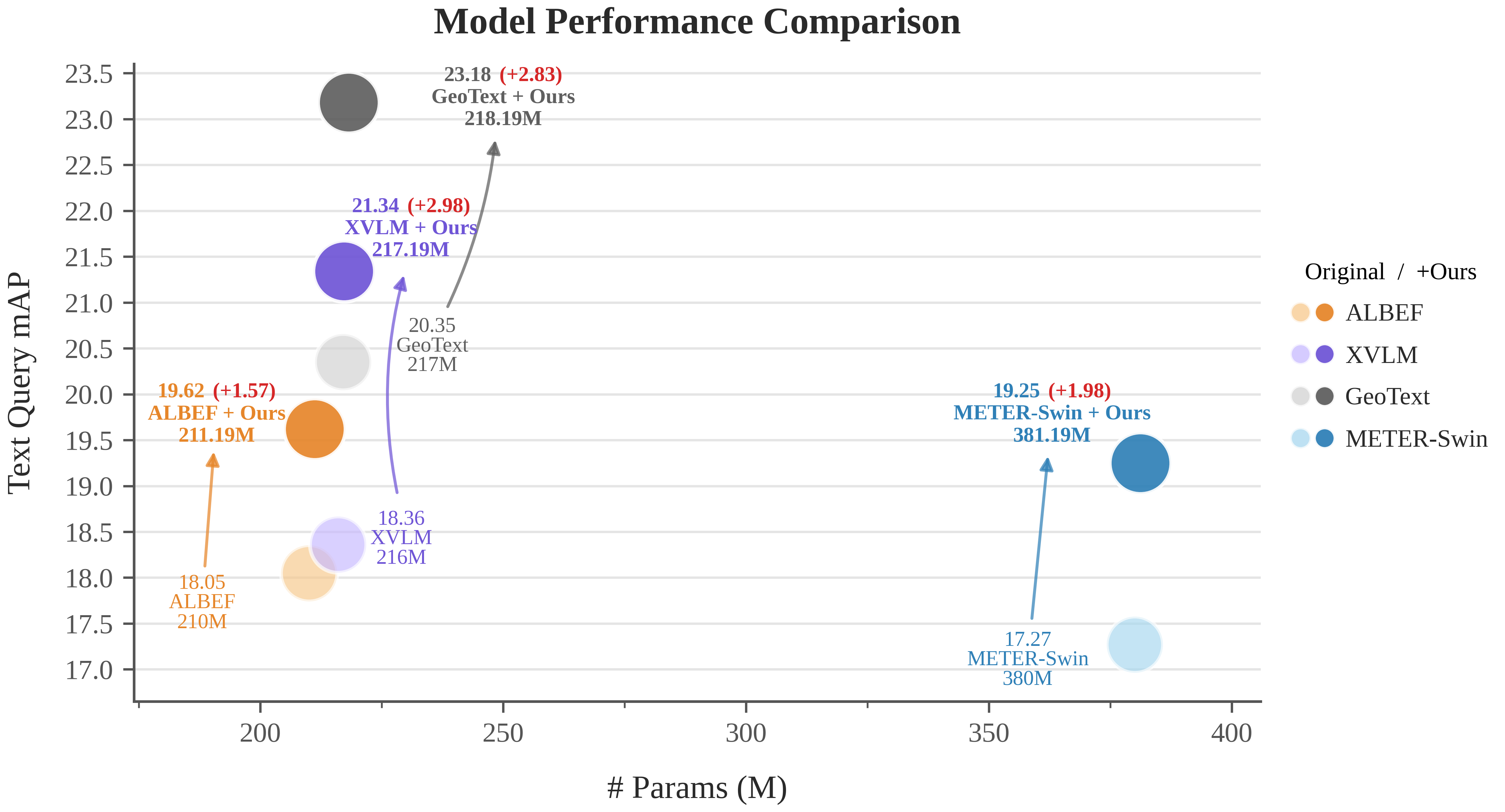}
    \caption{\textbf{Comparison of the proposed method with existing backbones in terms of Text Query mAP and parameter numbers on GeoText-1652.} We observe that our method brings consistent improvements over METER-Swin~\cite{dou2022empirical38}, ALBEF~\cite{li2021align36}, XVLM~\cite{zeng2021multi37}, and GeoText-1652~\cite{GeoText165206} while introducing only a marginal number of additional parameters.}
    \label{fig8}
    \vspace{-.20in}
\end{figure}

Recent advances in multi-modal large language models (MLLMs) have demonstrated strong potential in language understanding~\cite{feng2025urbanllava14,wu2025aeroduo15,liu2025can20,gong2025crossearth}, visual perception~\cite{he2024multi16,bigverdi2025perception17}, and cross-modal reasoning~\cite{liu2025aerialvg18,dong2025insight19}, suggesting a new modeling paradigm for addressing these challenges. 
We propose \textit{UniGeo}, a unified MLLM for text-guided drone geo-localization. 
As illustrated in Fig.~\ref{fig1}, UniGeo formulates a set of geospatial vision-language tasks within a shared modeling framework, including region grounding, image captioning, region-grounded description, image-text matching, pose-aware cross-view generation, and candidate verification. 
Although these tasks have different input-output forms, they are organized around three complementary capabilities required for text-guided geo-localization: geo-semantic understanding, cross-view semantic generation, and candidate-level fine-grained verification. 
In this way, UniGeo establishes a unified modeling process for query refinement, cross-view semantic reasoning, and discrimination among highly confusable candidates.
Specifically, UniGeo first establishes stable correspondences among local visual regions, spatial organization, and language descriptions through region-level semantic supervision and structured grounding, thereby providing a unified geo-semantic foundation for subsequent query enhancement and candidate discrimination.
To address the insufficiency of the original text query, UniGeo further performs candidate-conditioned geo-semantic generation and reformulation using the initially retrieved candidates, improving the completeness, stability, and discriminative power of the query representation. 
To tackle the difficulty of fine-grained disambiguation among highly similar candidates, UniGeo introduces a pose-aware cross-view generation branch that  constructs structure-preserving synthetic hard negatives under explicit geometric pose conditions (heading, pitch, altitude, and observation range). These synthesized samples are then used to train a lightweight verification head for re-scoring and fine-grained validation of highly confusable candidates. 
In addition, we design a three-stage training strategy that progressively organizes the learning of geo-semantic grounding, pose-aware cross-view generation, and candidate-level verification. Overall, UniGeo unifies these capabilities within a plug-and-play framework that remains seamlessly compatible with different retrieval backbones, without altering their original candidate generation mechanisms.
As shown in Fig.~\ref{fig8}, incorporating UniGeo consistently improves the Text Query mAP of representative retrieval backbones while introducing only marginal additional parameters.
The main contributions are as follows:
\begin{itemize}[label={},leftmargin=*,align=left]
     \item[$\bullet$] \textbf{Spatial-Reasoning Motivation}: We observe that hard-to-disambiguate text queries in drone geo-localization often require complex spatial reasoning rather than simple appearance matching. Resolving such queries requires joint understanding of linguistic intent and candidate-image spatial structure, motivating us to integrate region-level grounding, spatial-relation modeling, cross-view generation, and candidate verification within a unified MLLM.
     \item[$\bullet$] \textbf{Unified MLLM}: We propose UniGeo, a unified MLLM for text-guided drone geo-localization. Without modifying existing retrieval backbones, UniGeo integrates geo-semantic grounding, candidate-conditioned query enhancement, pose-aware cross-view generation, and lightweight candidate verification into a plug-and-play pipeline for fine-grained candidate disambiguation. A progressive three-stage training strategy further equips UniGeo with region-level semantic understanding, cross-view generative enhancement, and candidate-level discrimination, improving its ability to distinguish highly similar geographic candidates.
    \item[$\bullet$] \textbf{Competitive Geo-localization Performance}: The proposed method achieves consistent gains across diverse retrieval backbones under the text-guided geo-localization setting. Specifically, UniGeo achieves $45.37\%$ R@10 and $23.18\%$ mAP for text-guided geo-localization on GeoText-1652~\cite{GeoText165206}, improving the GeoText-1652 by $+ 13.59\%$ and $+ 2.83\%$, validating the effectiveness of candidate-conditioned reasoning and verification.
\end{itemize}



\section{Related Work}
\label{Related Work}

\noindent\textbf{Cross-view Geo-localization.} 
Cross-view geo-localization aims to establish correspondence between observations acquired from different platforms with substantial viewpoint variation~\cite{Zheng_Wei_Yang_202001,deuser2023sample4geo08,wang2024rethinking22,chen2024sdpl23,qian2026seeing}. Its central challenge lies in the representational inconsistency caused by large viewpoint changes, spatial layout distortion, and cross-platform imaging differences. Early studies mainly relied on hand-crafted local or global features combined with geometric verification for matching, but these approaches often struggled to remain robust in complex environments~\cite{zhang2021vector24,sanchez2013image25,barath2022learning26,shi2022accurate}. With the rise of deep learning, research gradually shifted toward more robust cross-view representation learning, improving alignment across viewpoints through metric learning~\cite{wang2024Muse07,vepa2024integrating48,choi2023depth49}, local region modeling~\cite{Wang_Zheng_Yan_Zhang_Sun_Zheng_Yang_202202,yang2021cross27,rodrigues2022global28,zhang2024geodtr+,wen2026fanet}, attention mechanisms~\cite{Lin_Zheng_Zhong_Luo_Li_Yang_Sebe_202203,yang2021cross27}, and Transformer-based interaction~\cite{Workman_Souvenir_Jacobs_201529,Lin_YinCui_Belongie_Hays_201530}. In the drone domain, the introduction of benchmarks such as University-1652 further advanced the development of drone-to-satellite geo-localization and encouraged subsequent methods to improve retrieval performance by exploiting finer-grained spatial structure and discriminative cues~\cite{Zheng_Wei_Yang_202001,shi2020optimal31,zhu2023sues32,ju2025video2bev45,ji2025game4loc46,dai2023vision47}.
Recent studies have extended cross-view geo-localization from image-only queries to text-based inputs. GeoText-1652~\cite{GeoText165206} first introduced natural language into drone geo-localization and established a benchmark for language-guided drone-to-satellite retrieval, while HCCM~\cite{ruan2025hccm33} and SAA-DGL~\cite{yuan2025seeing13} further improved performance through stronger alignment and transfer mechanisms. Although these methods (including those based on fine-grained vision-language models such as XVLM) perform cross-modal matching at the token or region level, they still primarily operate in the feature space and lack explicit geometric pose constraints for structure-preserving discrimination. In contrast, we focus on the post-retrieval stage and introduce pose-aware pixel-level generation, which imposes stricter spatial-position sensitivity and enables more reliable fine-grained verification among highly similar candidates.

\noindent\textbf{Vision-Language Models for Retrieval.}
In recent years, vision-language pretraining has substantially advanced image-text retrieval and matching, and has gradually become a key foundation for cross-modal geo-localization tasks~\cite{wen2025WeatherPrompt05,liu2025lamra54,tanaka2025vdocrag55,yu2025camel50,zeng2023x52,wang2023enhancing}.
Early large-scale contrastive pretraining methods, such as CLIP~\cite{radford2021learning35} and ALIGN~\cite{jia2021scaling51}, demonstrated that learning a shared embedding space from massive image-text pairs can yield highly transferable representations for zero-shot recognition and retrieval. 
Building on this line, subsequent studies further improved cross-modal retrieval from different perspectives, including align-before-fuse modeling in ALBEF~\cite{li2021align36}, fine-grained token-level interaction in FILIP~\cite{yao2021filip53}, multi-grained visual-semantic concept alignment in X-VLM~\cite{zeng2021multi37}, and end-to-end vision-language Transformer design in METER-Swin~\cite{dou2022empirical38}. 
Overall, this line of work mainly focuses on learning stronger shared representations and more effective matching backbones for direct image-text scoring. Different from these methods, we do not redesign the retrieval backbone itself. Instead, we take the initial candidate set returned by an existing retriever as input and perform post-retrieval reasoning and candidate verification on top of it.

\noindent\textbf{Unified Vision-Language Modeling.}
Beyond improving cross-modal retrieval backbones, recent studies have increasingly explored unified vision-language frameworks that jointly support retrieval, generation, and reasoning~\cite{feng2025urbanllava14,wu2025aeroduo15,wang2023image58,wang2022ofa57,wang2022align,gong2025crossearth}. Early efforts such as E2E-VLP~\cite{xu2021e2e39} already attempted to unify vision-language understanding and generation within a single framework. This direction was further advanced by Flamingo~\cite{alayrac2022flamingo40}, BLIP~\cite{li2022blip41}, and BLIP-2~\cite{li2023blip42}, which strengthened unified  understanding, generation, and reasoning over interleaved inputs. More recent studies moved closer to unified retrieval-generation-reasoning. FROMAGe~\cite{koh2023grounding43} jointly supports interleaved multimodal input processing, image retrieval, and text generation, while Sugar~\cite{chow2024unified44} further unifies retrieval, generation, and discriminative reasoning within the same multimodal system. Along this line, UniLIP~\cite{tang2025unilip59}, BAGEL~\cite{deng2025bagel}, and Qwen-Omni~\cite{qwen35omniblog} further extend unified multimodal modeling toward stronger understanding, generation, editing, and reasoning capabilities. Despite these advances, existing studies mainly target general-purpose multimodal tasks. In contrast, we focus on text-guided drone geo-localization, and build upon existing retrieval backbones to unify retrieval, generation, semantic refinement, and candidate verification within a reasoning pipeline for geo-localization.


\section{Method}
\label{Method}

\begin{figure*}[t]
    \centering
    \includegraphics[width=0.98\linewidth]{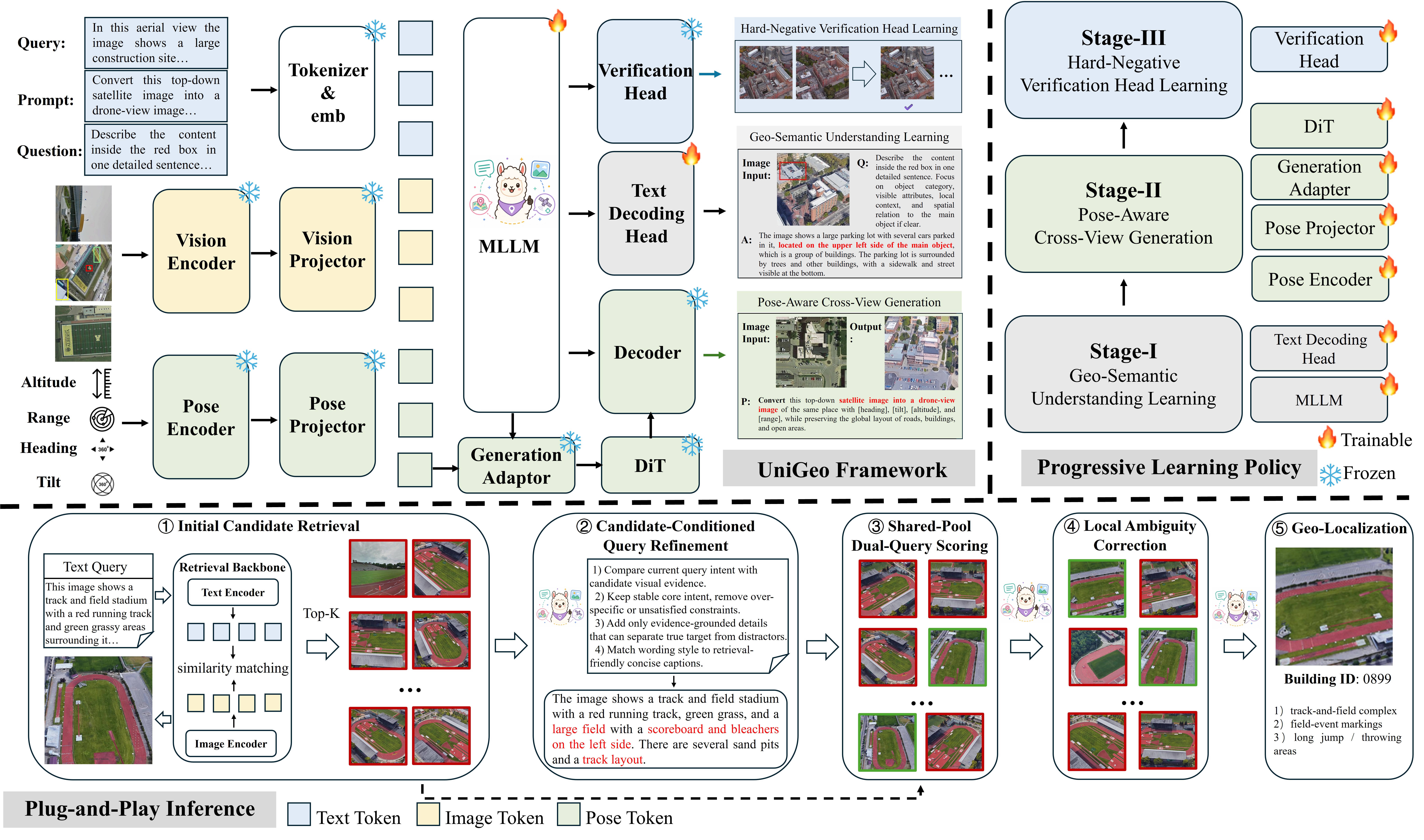}
    \caption{\textbf{The framework of UniGeo, including Progressive Learning Policy and Plug-and-Play Inference.} It first converts text, image, and pose inputs into unified tokens and processes them with a shared MLLM, together with task-specific heads for text decoding, pose-aware generation, and candidate verification. The model is optimized through a progressive three-stage learning policy: Stage-I learns geo-semantic understanding with region-level grounding and spatial-relation supervision, Stage-II learns pose-aware cross-view generation under explicit geometric conditions, and Stage-III trains a hard-negative verification head for fine-grained candidate discrimination. During plug-and-play inference, UniGeo operates on the candidate pool returned by an external retrieval backbone, performs candidate-conditioned query refinement and candidate-level verification, and finally re-ranks highly confusable candidates for more reliable geo-localization.}
    \label{fig2}
    \vspace{-.20in}
\end{figure*}

As shown in Fig.~\ref{fig2}, UniGeo is designed as a unified MLLM framework that integrates understanding, training, and inference. \textbf{UniGeo Framework} defines the shared semantic foundation by learning a unified geospatial representation space, which supports local semantic understanding, explicit language-to-region correspondence, and structured spatial-relation modeling. \textbf{Progressive Learning Policy} then acquires these capabilities through a progressive multi-stage strategy, where geospatial semantic grounding, cross-view semantic augmentation, and candidate-level verification are optimized in sequence. \textbf{Plug-and-Play Inference} finally deploys the learned reasoning capability in a plug-and-play manner: given the candidate pool produced by an arbitrary retrieval backbone, it performs query refinement, candidate verification, and local reranking to produce the final geo-localization result.

\subsection{UniGeo Framework}
Building on the success of MLLMs across a wide range of domains, we choose UniLIP~\cite{tang2025unilip59} as the base model for our UniGeo experiments. 
Based on its shared vision-language backbone, we construct a unified geo-semantic representation space. Rather than relying solely on holistic image captioning, we further emphasize the correspondence between language and local visual regions and enhance model geo-semantic understanding through two complementary forms of supervision: region-level semantic supervision and structured discriminative supervision.

\textbf{Region-level Semantic Supervision.}
Given an image $I$ and its region-level annotation set $\{(b_m, y_m)\}_{m=1}^{M}$, where $b_m$ denotes the bounding box of the $m$-th local region and $y_m$ denotes its associated textual description, we construct each valid bbox-sentence pair as a region-level supervised sample. Unlike whole-image captioning, the objective here is to generate the semantic description of a specified local region under explicit region conditioning.
We adopt a hybrid region prompt that combines discretized bounding-box coordinate tokens with explicit textual region indicators. To explicitly specify the target region in the visual input, we use the full image with bbox overlay and, when applicable, the corresponding ROI crop as additional local visual evidence.
Under such region-aware conditions, the model is trained to autoregressively generate the target region description $y_m$. We employ a standard conditional language modeling objective and compute the loss only over the answer tokens:
\begin{equation}
\label{eq1}
\mathcal{L}_{\mathrm{reg}}
=
-\sum_m \sum_{t\in A_m}
\log P_{\theta}\!\left(y_{m,t}\mid y_{m,<t}, I, b_m\right),
\end{equation}
where $A_m$ denotes the index set of answer tokens involved in loss computation, and \(P_{\theta}\) denotes the autoregressive token probability predicted by the model parameterized by \(\theta\).
This supervision establishes stable correspondences between local visual regions and language descriptions.

\textbf{Structured Discriminative Supervision.}
Building upon the local semantic correspondences established by region-level semantic supervision, we further introduce structured discriminative supervision to explicitly model both the grounding relationship between local descriptions and visual regions, and the spatial organization among multiple regions.
For each valid region-sentence pair $(b_m, y_m)$, we first tokenize the regional description $y_m$ into a textual token sequence $T_m$. 
A language-to-region geo-localization head then models the interaction between $T_m$ and the full-image token sequence $V$ and predicts the referred bounding box $\hat{b}_m$. 
The geo-localization loss is defined as:
\begin{equation}
\label{eq2}
\mathcal{L}_{\mathrm{bbox}} 
= 
\|\hat{b}_m - b_m\|_1 
+ 
1 - \mathrm{GIoU}(\hat{b}_m, b_m),
\end{equation}
where $b_m$ and $\hat{b}_m$ denote the ground-truth and predicted bounding boxes, respectively, and $\mathrm{GIoU}(\cdot,\cdot)$ denotes the generalized intersection-over-union.
This supervision explicitly enforces the grounding correspondence between each local description and its target visual region.
We further impose spatial relation supervision over any valid region pair $(b_m, b_n)$ within the same image. 
Specifically, we reshape the shared image token sequence $V$ into a two-dimensional feature map and extract region representations $r_m$ and $r_n$ from the ground-truth boxes via RoIAlign. 
A spatial relation head then takes the paired region representations and their box geometry as input, and predicts the horizontal and vertical relation logits $(\hat{r}^{h}_{mn}, \hat{r}^{v}_{mn})$ between the two regions. 
The horizontal relation categories are defined as \textit{left / center / right}, and the vertical relation categories are defined as \textit{upper / middle / lower}.
During training, we apply classification losses to the two directions separately and sum them to obtain the spatial relation supervision:
\begin{equation}
\label{eq4}
\mathcal{L}_{\mathrm{spa}} = \mathcal{L}_h(\hat{r}_{mn}^{h}, r_{mn}^{h}) + \mathcal{L}_v(\hat{r}_{mn}^{v}, r_{mn}^{v}),
\end{equation}
where $r_{mn}^{h}$ and $r_{mn}^{v}$ denote the ground-truth horizontal and vertical relation labels, and $\mathcal{L}_{h}$ and $\mathcal{L}_{v}$ are cross-entropy losses for horizontal and vertical spatial-relation classification.

\subsection{Progressive Learning Policy}
UniGeo is trained with a progressive multi-stage strategy that stages the acquisition of different capabilities within a unified model.
In Stage I, region-level semantic supervision and structured discriminative supervision are introduced to establish stable correspondences among local regions, spatial structure, and language descriptions.
In Stage II, a pose-aware cross-view generation mechanism is further incorporated to learn structure-preserving semantic transformation under explicit geometric constraints, while constructing synthetic samples to enhance cross-view modeling.
In Stage III, a candidate-level verification head is trained using both real and synthetic hard negatives, thereby improving fine-grained disambiguation among highly similar candidates.

\textbf{Stage-I: Geo-Semantic Understanding Learning.}
The objective is to establish stable correspondences among local regional semantics, explicit language-to-region grounding, and inter-region spatial structure within the shared vision-language backbone. To this end, we jointly optimize the model with region-level semantic supervision, global image-text alignment, fine-grained image-text matching, language-to-region geo-localization supervision, and inter-region spatial relation supervision. The overall objective of this stage is formulated as:
\begin{equation}
\label{eq5}
\mathcal{L}_{\text{Stage-I}}
=
\mathcal{L}_{\mathrm{reg}}
+ \mathcal{L}_{\mathrm{itc}}
+ \mathcal{L}_{\mathrm{itm}}
+ \mathcal{L}_{\mathrm{spa}}
+ 0.1\, \mathcal{L}_{\mathrm{bbox}},
\end{equation}
where $\mathcal{L}_{\mathrm{reg}}$ encourages the model to learn region-level semantic expressions, $\mathcal{L}_{\mathrm{itc}}$ and $\mathcal{L}_{\mathrm{itm}}$ respectively enforce global image-text alignment and fine-grained image-text matching, $\mathcal{L}_{\mathrm{spa}}$ constrains the relative spatial organization among regions, and $\mathcal{L}_{\mathrm{bbox}}$ establishes explicit grounding from linguistic descriptions to visual regions.
Following existing works~\cite{GeoText165206}, we set the coefficient of $\mathcal{L}_{\mathrm{bbox}}$ to 0.1 for loss balancing.
We freeze the diffusion generation branch and optimize only the shared vision-language backbone together with the associated grounding and discriminative modules, thereby establishing a stable foundation for subsequent cross-view modeling and discrimination.

\textbf{Stage-II: Pose-Aware Cross-View Generation.}
After establishing a stable geo-semantic understanding foundation in Stage-I, we further introduce pose-aware cross-view generative learning to model the semantic and geometric consistency of the same target scene across different observation viewpoints. Specifically, we adopt a latent-diffusion-based generation framework, where a Diffusion Transformer (DiT)~\cite{peebles2023scalable} serves as the core generative backbone for learning structure-preserving view transformation between satellite and drone perspectives. To explicitly inject viewpoint information, we encode pose attributes, including heading, pitch, altitude, and observation range, into pose conditions to guide the diffusion generation process. The pose metadata are used as diffusion conditions, while the teacher-based pose-consistency loss is applied to heading and observation range.

Stage-II inherits the model parameters from Stage-I and freezes the shared vision-language understanding backbone, including the visual encoder, language model, and cross-modal connector, so as to preserve the geo-semantic correspondences learned in Stage-I. 
To provide explicit geometric supervision for generated samples, we further introduce a pose teacher module. This module consists of a frozen visual encoder and two lightweight regression heads, which estimate the heading angle and observation range of the input image, respectively. During cross-view generative learning, the pose teacher remains frozen and is used only as an external supervision signal to impose pose-consistency constraints on generated samples.
In our final instantiation, the training objective of this stage is defined as:
\begin{equation} 
\label{eq6} 
\mathcal{L}_{\text{Stage-II}} 
= 
\mathcal{L}_{\mathrm{diff}} 
+ 
\lambda_p\mathcal{L}_{\mathrm{pose}} 
+ 
\lambda_h\mathcal{L}_{\mathrm{heading}}^{\mathrm{aux}}, 
\end{equation} 
where $\mathcal{L}_{\mathrm{diff}}$ denotes the standard diffusion denoising loss, $\mathcal{L}_{\mathrm{pose}}$ denotes the teacher-based pose-consistency loss imposed on generated samples, and $\mathcal{L}_{\mathrm{heading}}^{\mathrm{aux}}$ denotes an auxiliary heading regularization term applied to the pose-conditioned hidden representation. 
The teacher-based pose-consistency loss is decomposed into heading and range consistency terms: \begin{equation} 
\label{eq7} 
\mathcal{L}_{\mathrm{pose}} 
= \mathcal{L}_{\mathrm{heading}}^{\mathrm{teacher}} 
+ \mathcal{L}_{\mathrm{range}}^{\mathrm{teacher}}.
\end{equation}
The heading term penalizes the discrepancy between the teacher-predicted heading vector and the sine-cosine encoding of the ground-truth heading angle, while the range term uses a Smooth L1 loss between the teacher-predicted observation range and the normalized range target.
In addition, we introduce an auxiliary heading regularization term on the pose-conditioned hidden representation: 
\begin{equation} 
\label{eq10} 
\mathcal{L}_{\mathrm{heading}}^{\mathrm{aux}} 
= 
\frac{1}{|\mathcal{D}_{\mathrm{pose}}|} 
\sum_{i\in\mathcal{D}_{\mathrm{pose}}} 
\left\| 
\tilde{h}_i 
- 
\begin{bmatrix} 
\cos\theta_i \\ 
\sin\theta_i 
\end{bmatrix} 
\right\|_2^2, 
\end{equation}
where $\mathcal{D}_{\mathrm{pose}}$ denotes the set of training samples with valid pose annotations, $\theta_i$ is the ground-truth heading angle, and $\tilde{h}_i$ is the heading vector predicted from the pose-conditioned hidden representation.

\textbf{Stage-III: Hard-Negative Verification Head Learning.}
We train a lightweight candidate verification head in Stage-III to improve fine-grained discrimination among highly confusable candidates. 
Specifically, we freeze both the shared backbone and the generation branch, and optimize only the parameters of the verification head on top of the query-candidate fused representations extracted by the frozen backbone. 
For each text query $q$, we construct a training group consisting of one positive sample $I^{+}$, a set of real hard negatives $I_{r}^{-}$, and a set of synthetic hard negatives $I_{s}^{-}$. 
The real hard negatives are mined from retrieved candidates, while the synthetic hard negatives are produced by the cross-view generation branch learned in Stage-II. 
The frozen backbone extracts query-candidate fused representations $\Phi(q,I)$, which are fed into a lightweight verification head $g(\cdot)$ to produce matching scores. 
The verification objective is defined as:
\begin{equation}
\label{eq11}
\mathcal{L}_{\text{Stage-III}} 
= 
\mathcal{L}_{\mathrm{real}} 
+ 
\lambda_{\mathrm{syn}} \mathcal{L}_{\mathrm{syn}},
\end{equation}
where $\mathcal{L}_{\mathrm{real}}$ is the binary matching loss on real hard negatives, and $\mathcal{L}_{\mathrm{syn}}$ is an auxiliary binary matching loss on synthetic hard negatives produced by the cross-view generation branch.

\begin{equation}
\label{eq12}
\mathcal{L}_{\mathrm{real}}
=
\mathcal{L}_{\mathrm{BCE}}\!\bigl(g(\Phi(q,I^{+})),1\bigr)
+
\sum_{I_{r}^{-}}
\mathcal{L}_{\mathrm{BCE}}\!\bigl(g(\Phi(q,I_{r}^{-})),0\bigr),
\end{equation}

\begin{equation}
\label{eq13}
\mathcal{L}_{\mathrm{syn}}
=
\sum_{I_{s}^{-}}
\mathcal{L}_{\mathrm{BCE}}\!\bigl(g(\Phi(q,I_{s}^{-})),0\bigr),
\end{equation}
where $\mathcal{L}_{\mathrm{BCE}}$ denotes the binary cross-entropy loss, with label $1$ for the positive query-candidate pair and label $0$ for hard negative pairs.

Overall, Stage-III enhances candidate-level discrimination by integrating Stage-I grounding and Stage-II cross-view generation through hard-negative verification learning.

\subsection{Plug-and-Play Inference}
At inference time, UniGeo is deployed in a plug-and-play manner on top of arbitrary external retrieval backbones, without altering their original training protocol or initial retrieval pipeline.

\textbf{Initial Candidate Retrieval.}
Given a text query $q$ and an image gallery $\mathcal{G}$, an external retrieval backbone $B$ first performs high-recall candidate generation and returns the top-\(K_0\) candidates as the initial candidate pool \(C_0(q)\), where \(K_0\) denotes the pool size. Rather than redesigning this stage, we treat it as a candidate generation step and perform subsequent fine-grained reasoning on top of the retrieved candidates.

\textbf{Candidate-Conditioned Query Refinement.}
To compensate for the incompleteness of the original text query, we generate a refined query $\tilde{q}$ conditioned on both the original query and the initial candidate context:
\(\tilde{q} = R_{\phi}(q, C_0(q))\), where $R_{\phi}(\cdot)$ denotes the candidate-conditioned query refinement function. This process is not intended as free-form rewriting, but aims to supplement localization-relevant semantic cues under the constraint of the retrieved candidate context. The candidates retrieved by the refined query are denoted as $C_{\mathrm{ref}}(\tilde{q})$, and the shared candidate pool is constructed as: \(C(q) = C_0(q) \cup C_{\mathrm{ref}}(\tilde{q})\).

\textbf{Shared-Pool Dual-Query Scoring.}
For each candidate image $I_i \in \mathcal{C}(q)$, we compute its candidate-level matching scores with respect to the original query $q$ and the refined query $\tilde q$:
\begin{equation}
\label{eq15}
s_i^{(o)} = M(q,I_i),
\qquad s_i^{(r)} = M(\tilde{q},I_i),
\end{equation}
We estimate a lightweight refinement reliability score from the semantic consistency between $q$ and $\tilde q$, the overlap between $C_0(q)$ and $C_{\mathrm{ref}}(\tilde q)$, and the amount of newly introduced semantic content. The normalized reliability score is mapped to an adaptive refinement weight $\beta_q$, and the two scores are combined as:
\begin{equation}
\label{eq16}
s_i = (1-\beta_q)s_i^{(o)} + \beta_q s_i^{(r)},
\end{equation}
where $s_i$ denotes the gated dual-query candidate score.

\textbf{Local Ambiguity Correction and Geo-Localization.}
Building on the primary verification score $s_i$, we further select the top-$K$ candidates as a local shortlist. For candidates within the shortlist, the lightweight verification head trained in Stage-III produces an additional candidate-level score $s_i^h$, which is fused with $s_i$ in probability space:
\(\hat{s}_i=(1-\alpha)s_i+\alpha s_i^h\), where $\alpha$ denotes the fusion weight of the verification head score. For candidates outside the shortlist, their scores remain unchanged. Finally, we rerank the candidates according to the corrected scores and select the highest-scoring candidate as the geo-localization result.


\section{Experiment}
\label{experiment}

\subsection{Implementation Details}
\textbf{MLLMs.} Our method is complementary to existing retrieval approaches and can be seamlessly integrated with different backbones. In our implementation, we adopt UniLIP~\cite{tang2025unilip59} as the underlying MLLM backbone of UniGeo and conduct all experiments on 4$\times$NVIDIA RTX A6000 GPUs. The shared vision-language backbone is initialized from UniLIP and paired with the InternVL3-2B~\cite{zhu2025internvl3} processor. At the framework level, UniGeo can be plugged into arbitrary external candidate-retrieval backbones.

\textbf{Training.} Training follows a three-stage progressive strategy. 
In Stage-I, we conduct region-level VQA fine-tuning on GeoText-1652\cite{GeoText165206} using the \texttt{bbox\_box} prompt format and red-box region annotations, updating only the last eight layers of the language model, followed by multi-task foundation training. The VQA fine-tuning stage runs for 1 epoch with a learning rate of $5\times10^{-6}$, a maximum sequence length of 1024, a per-device batch size of 1, and 8 gradient accumulation steps. The foundation training stage also runs for 1 epoch, with a batch size of 32, a maximum text length of 50, a learning rate of $1\times10^{-4}$.
In Stage II, we train the pose-aware cross-view generation branch on University-1652~\cite{Zheng_Wei_Yang_202001} using bidirectional drone-satellite editing samples with pose metadata, together with generative image-text samples. This stage starts with 1000 pose warm-up steps, followed by partial unfreezing of the last four DiT blocks for 5 additional epochs. Training uses bf16 precision, per-device batch size 128, learning rate $1\times10^{-4}$, maximum sequence length 1024, and a cosine scheduler with minimum learning rate $1\times10^{-5}$. 
In Stage III, the lightweight verification head is initialized from the original ITM head, and each query is paired by default with two real hard negatives and two synthetic hard negatives. This stage is trained for 1 epoch with batch size 8, image resolution 384, maximum text length 50, a learning rate of $1\times10^{-4}$, weight decay of $1\times10^{-2}$, \(\lambda_{\mathrm{syn}}\) set to 0.3, \(\lambda_{\mathrm{p}}\) set to 0.2 and \(\lambda_{\mathrm{h}}\) set to 0.1. 

\textbf{Inference.} At inference time, the external retrieval backbone first returns the top-$K_0$ candidates, where $K_0=256$. The original-query and refined-query scores are fused by the lightweight gating mechanism, with the adaptive refinement weight $\beta_q \in [0.12,0.55]$. We then select the top-$K$ candidates, with ($K=16$), as the local shortlist for Stage-III verification. The verification head scores are fused with the primary verification scores in probability space, with $\alpha$ set to 0.1.

\textbf{Evaluation Metrics.}
We evaluate UniGeo from three perspectives: geo-localization, text generation, and cross-view image generation. For geo-localization, we report Recall@1, Recall@5, Recall@10 and mAP as the primary metrics. For text generation, we use BLEU-1, BLEU-2, and BLEU-4, while for image generation, we report FID, SSIM, PSNR, and pose/heading control accuracy to assess both visual quality and geometric controllability. Text and image generation metrics are used as auxiliary indicators to support the analysis of the main geo-localization results.

\textbf{Reproducibility.} The code and model weights will be publicly released.

\subsection{Main Results}
We conduct experiments to evaluate UniGeo from three complementary perspectives, corresponding to its retrieval, query refinement, and cross-view generation capabilities. First, we report the overall performance on geo-localization to validate the effectiveness of UniGeo as a plug-and-play post-retrieval reasoning framework. Second, we assess the quality of candidate-conditioned query refinement, aiming to determine whether the unified vision-language backbone can provide discriminative textual evidence for candidate-level ambiguity resolution. Third, we evaluate pose-aware cross-view image generation to examine whether UniGeo captures viewpoint-dependent geometric variations and produces visually consistent cross-view transformations.

\begin{table*}[t]
\centering
\scriptsize
\setlength{\tabcolsep}{4pt}
\renewcommand{\arraystretch}{1.08}
\caption{\textbf{Re-implemented image-text bidirectional retrieval results on GeoText-1652~\cite{GeoText165206}.}
The first block reports direct evaluation results of pre-trained models, the second block reports results after fine-tuning on GeoText-1652, and the last block compares recent geo-localization methods. 
Text Query denotes text-to-image search, and Image Query denotes image-to-text search. 
We report Recall@K and mAP as evaluation metrics.}
\resizebox{\textwidth}{!}{%
\begin{tabular}{l|cccc|cccc}
\toprule
\multirow{2}{*}{Method} 
& \multicolumn{4}{c|}{Text Query} 
& \multicolumn{4}{c}{Image Query} \\
 & R@1 & R@5 & R@10 & mAP & R@1 & R@5 & R@10 & mAP \\
\hline

METER-Swin~\cite{dou2022empirical38} 
& 0.98 & 2.23 & 3.07 & 1.61  
& 1.43 & 3.95 & 5.81 & 2.54 \\
\rowcolor{gray!15}
METER-Swin + Ours  
& 1.03\gain{+0.05} & 2.29\gain{+0.06} & 3.18\gain{+0.11} & 1.61\same{+0.00}
& 1.43\same{+0.00} & 4.04\gain{+0.09} & 5.90\gain{+0.09} & 2.62\gain{+0.08} \\

ALBEF-4M~\cite{li2021align36} 
& 1.81 & 4.56 & 6.59 & 3.23
& 2.21 & 7.37 & 11.51 & 4.68 \\
\rowcolor{gray!15}
ALBEF-4M + Ours  
& 1.82\gain{+0.01} & 5.04\gain{+0.48} & 7.38\gain{+0.79} & 3.37\gain{+0.14}
& 3.19\gain{+0.98} & 9.56\gain{+2.19} & 13.61\gain{+2.10} & 6.12\gain{+1.44} \\

ALBEF-14M~\cite{li2021align36} 
& 1.84 & 4.83 & 6.69 & 3.34
& 4.59 & 13.12 & 18.95 & 8.69 \\
\rowcolor{gray!15}
ALBEF-14M + Ours  
& 1.96\gain{+0.12} & 5.20\gain{+0.37} & 7.11\gain{+0.42} & 3.40\gain{+0.06}
& 5.92\gain{+1.33} & 15.13\gain{+2.01} & 20.71\gain{+1.76} & 10.24\gain{+1.55} \\

XVLM-4M~\cite{zeng2021multi37} 
& 4.71 & 9.85 & 13.30 & 7.54 
& 5.53 & 14.64 & 21.53 & 10.00 \\
\rowcolor{gray!15}
XVLM-4M + Ours 
& 4.97\gain{+0.26} & 10.42\gain{+0.57} & \second{14.31}\gain{+1.01} & 7.84\gain{+0.30}
& 7.43\gain{+1.90} & 17.98\gain{+3.34} & 24.24\gain{+2.71} & 12.34\gain{+2.34} \\

XVLM-16M~\cite{zeng2021multi37} 
& \second{5.27} & \second{10.71} & 14.29 & \second{8.21} 
& \second{7.50} & \second{20.29} & \second{28.90} & \second{13.64} \\
\rowcolor{gray!15}
XVLM-16M + Ours 
& \best{5.50}\gain{+0.23} & \best{13.27}\gain{+2.56} & \best{18.40}\gain{+4.11} & \best{9.04}\gain{+0.83}
& \best{9.12}\gain{+1.62} & \best{23.00}\gain{+2.71} & \best{31.13}\gain{+2.23} & \best{15.58}\gain{+1.94} \\

\hline

METER-Swin$_{(\mathit{finetuned})}$ 
& 12.16 & 20.65 & 25.92 & 17.27 
& 23.10 & 49.08 & 62.47 & 34.33 \\
\rowcolor{gray!15}
METER-Swin + Ours 
& 12.48\gain{+0.32} & 27.42\gain{+6.77} & 35.45\gain{+9.53} & 19.25\gain{+1.98}
& 24.25\gain{+1.15} & 50.75\gain{+1.67} & 63.23\gain{+0.76} & 36.53\gain{+2.20} \\

ALBEF-4M$_{(\mathit{finetuned})}$ 
& 12.24 & 20.35 & 25.29 & 17.14 
& 21.91 & 46.77 & 59.28 & 32.83 \\
\rowcolor{gray!15}
ALBEF-4M + Ours  
& 12.61\gain{+0.37} & 25.45\gain{+5.10} & 32.55\gain{+7.26} & 18.73\gain{+1.59}
& 23.27\gain{+1.36} & 47.96\gain{+1.19} & 60.32\gain{+1.04} & 34.92\gain{+2.09} \\

ALBEF-14M$_{(\mathit{finetuned})}$ 
& 12.96 & 21.43 & 26.43 & 18.05 
& 26.61 & 54.09 & 66.71 & 38.28 \\
\rowcolor{gray!15}
ALBEF-14M + Ours 
& \best{13.46}\gain{+0.50} & 25.98\gain{+4.55} & 33.36\gain{+6.93} & 19.62\gain{+1.57}
& \second{27.10}\gain{+0.49} & \second{54.34}\gain{+0.25} & \second{66.72}\gain{+0.01} & \second{39.82}\gain{+1.54} \\

XVLM-4M$_{(\mathit{finetuned})}$ 
& 12.47 & 21.80 & 27.48 & 17.95 
& 25.78 & 53.47 & 66.30 & 38.81 \\
\rowcolor{gray!15}
XVLM-4M + Ours 
& 13.08\gain{+0.61} & \second{31.40}\gain{+9.60} & \best{42.84}\gain{+15.36} & \second{21.10}\gain{+3.15}
& 26.60\gain{+0.82} & 54.18\gain{+0.71} & \best{66.80}\gain{+0.50} & 39.47\gain{+0.66} \\

XVLM-16M$_{(\mathit{finetuned})}$ 
& 12.90 & 22.17 & 27.80 & 18.36 
& 27.02 & 54.02 & 66.30 & 39.69 \\
\rowcolor{gray!15}
XVLM-16M + Ours  
& \second{13.34}\gain{+0.44} & \best{31.62}\gain{+9.45} & \second{42.82}\gain{+15.02} & \best{21.34}\gain{+2.98}
& \best{27.11}\gain{+0.09} & \best{54.38}\gain{+0.36} & 66.63\gain{+0.33} & \best{39.88}\gain{+0.19} \\

\hline

GeoText-1652~\cite{GeoText165206}
& 13.97 & 25.20 & 31.78 & 20.35 
& 28.81 & 56.45 & 69.30 & 40.60 \\

\rowcolor{gray!15}
GeoText-1652 + Ours 
& 14.72\gain{+0.75} & \second{33.80}\gain{+8.60} & \best{45.37}\gain{+13.59} & 23.18\gain{+2.83}
& \best{29.32}\gain{+0.51} & 56.80\gain{+0.35} & \second{69.57}\gain{+0.27} & \best{41.98}\gain{+1.38} \\

HCCM~\cite{ruan2025hccm33}
& 14.51 & 25.93 & 32.64 & 21.07 
& 28.66 & \second{57.10} & 69.43 & 41.38 \\

\rowcolor{gray!15}
HCCM + Ours 
& 15.05\gain{+0.54} & \best{34.21}\gain{+8.28} & \second{45.23}\gain{+12.59} & \second{23.57}\gain{+2.50}
& \second{28.93}\gain{+0.27} & \best{57.52}\gain{+0.42} & \best{69.96}\gain{+0.53} & \second{41.69}\gain{+0.31} \\

NGCG-MLLMs~\cite{chen2026turning}
& \second{16.15} & 29.69 & 37.55 & 23.43 
& 16.42 & 38.04 & 50.10 & 26.47 \\

\rowcolor{gray!15}
NGCG-MLLMs + Ours 
& \best{17.52}\gain{+1.37} & 31.75\gain{+2.06} & 39.59\gain{+2.04} & \best{24.74}\gain{+1.31}
& 16.56\gain{+0.14} & 38.81\gain{+0.77} & 50.69\gain{+0.59} & 26.61\gain{+0.14} \\

\bottomrule
\end{tabular}%
}
\label{tab1}
\vspace{-0.20in}
\end{table*}

\textbf{Geo-Localization Performance.}
As shown in Tables~\ref{tab1} and~\ref{tab100}, we report the overall geo-localization performance of UniGeo on GeoText-1652~\cite{GeoText165206} and UAVReason~\cite{sun2026uavreason}. In addition to GeoText-1652, we further construct an image-text geo-localization benchmark based on UAVReason. We retain its original scene-level training/test split and use Qwen3.6-27B~\cite{qwen3.6-27b} to generate five natural-language scene descriptions for each UAV image, with the training and test sets remaining completely disjoint at the scene level.
Since the images in UAVReason are collected along continuous UAV trajectories, adjacent frames exhibit strong continuity in both geographic location and spatial coverage. We therefore adopt a temporal-neighborhood-based localization evaluation protocol. Specifically, for each query image, images within the preceding and following five frames in the same scene are all regarded as correct localization results. For image-to-text retrieval, the natural-language descriptions associated with these neighboring images are likewise treated as positives. 

The text-query setting serves as the primary evaluation protocol, where the goal is to retrieve and localize the target region from natural-language descriptions. The image-query setting is included as an auxiliary protocol to examine whether the geo-semantic discrimination capability learned by UniGeo can transfer to another query modality. Overall, UniGeo consistently improves the retrieval performance of a wide range of representative methods on both datasets, including pre-trained and fine-tuned vision-language retrieval backbones such as METER-Swin~\cite{dou2022empirical38}, ALBEF~\cite{li2021align36}, and XVLM~\cite{zeng2021multi37}, as well as recent geo-localization methods such as GeoText-1652~\cite{GeoText165206}, HCCM~\cite{ruan2025hccm33}, and NGCG-MLLMs~\cite{chen2026turning}.
These results demonstrate that UniGeo can be effectively adapted to different datasets, retrieval backbones, and geo-localization pipelines, validating its effectiveness and generalizability as a plug-and-play post-retrieval reasoning framework.


\begin{table*}[t]

\centering

\scriptsize

\setlength{\tabcolsep}{4pt}

\renewcommand{\arraystretch}{1.08}

\caption{\textbf{Re-implemented image-text bidirectional retrieval results on UAVReason~\cite{sun2026uavreason}.}
The first block reports direct evaluation results of pre-trained models, the second block reports results after fine-tuning on UAVReason, and the last block compares recent geo-localization methods.
Text Query denotes text-to-image search, and Image Query denotes image-to-text search.
We report Recall@K and mAP as evaluation metrics.}

\resizebox{\textwidth}{!}{%

\begin{tabular}{l|cccc|cccc}

\toprule

\multirow{2}{*}{Method}
& \multicolumn{4}{c|}{Text Query}
& \multicolumn{4}{c}{Image Query} \\

& R@1 & R@5 & R@10 & mAP
& R@1 & R@5 & R@10 & mAP \\

\hline

METER-Swin~\cite{dou2022empirical38}
& 2.71 & 6.73 & 8.66 & 4.44
& 3.48 & 10.32 & 14.89 & 1.04 \\

\rowcolor{gray!15}
METER-Swin + Ours
& 2.74\gain{+0.03} & 7.24\gain{+0.51} & 9.30\gain{+0.64} & 4.57\gain{+0.13}
& 3.61\gain{+0.13} & 10.56\gain{+0.24} & 15.07\gain{+0.18} & 1.12\gain{+0.08} \\

ALBEF-4M~\cite{li2021align36}
& 4.44 & 11.25 & 15.65 & 7.39
& 4.76 & 15.17 & 24.93 & 1.81 \\

\rowcolor{gray!15}
ALBEF-4M + Ours
& 4.50\gain{+0.06} & 12.68\gain{+1.43} & 17.48\gain{+1.83} & 7.89\gain{+0.50}
& 4.84\gain{+0.08} & 16.92\gain{+1.75} & 25.71\gain{+0.78} & 1.97\gain{+0.16} \\

ALBEF-14M~\cite{li2021align36}
& 5.35 & 12.91 & 17.97 & 8.72
& 5.37 & 17.51 & 26.52 & 2.13 \\

\rowcolor{gray!15}
ALBEF-14M + Ours
& 5.40\gain{+0.05} & 13.52\gain{+0.61} & 18.29\gain{+0.32} & 8.87\gain{+0.15}
& \second{5.60}\gain{+0.23} & 18.20\gain{+0.69} & 27.17\gain{+0.65} & 2.21\gain{+0.08} \\

XVLM-4M~\cite{zeng2021multi37}
& 8.60 & 20.98 & 29.66 & 14.00
& 4.70 & 15.86 & 25.25 & 1.83 \\

\rowcolor{gray!15}
XVLM-4M + Ours
& 8.71\gain{+0.11} & 21.52\gain{+0.54} & \second{30.70}\gain{+1.04} & 14.13\gain{+0.13}
& 4.70\same{+0.00} & \second{19.14}\gain{+3.28} & \second{31.22}\gain{+5.97} & \second{2.31}\gain{+0.48} \\

XVLM-16M~\cite{zeng2021multi37}
& \second{9.27} & \second{21.66} & 30.18 & \second{14.77}
& \second{5.60} & 16.60 & 27.28 & 2.10 \\

\rowcolor{gray!15}
XVLM-16M + Ours
& \best{9.51}\gain{+0.24} & \best{22.68}\gain{+1.02} & \best{30.98}\gain{+0.80} & \best{15.16}\gain{+0.39}
& \best{5.70}\gain{+0.10} & \best{20.95}\gain{+4.35} & \best{32.90}\gain{+5.62} & \best{2.38}\gain{+0.28} \\

\hline

METER-Swin$_{(\mathit{finetuned})}$
& 13.42 & 25.70 & 33.62 & 18.77
& 6.00 & 12.48 & 15.76 & 2.07 \\

\rowcolor{gray!15}
METER-Swin + Ours
& 13.45\gain{+0.03} & 26.73\gain{+1.03} & 34.51\gain{+0.89} & 18.96\gain{+0.19}
& 6.21\gain{+0.21} & 12.76\gain{+0.28} & 15.98\gain{+0.22} & 2.40\gain{+0.33} \\

ALBEF-4M$_{(\mathit{finetuned})}$
& 7.73 & 17.05 & 22.67 & 11.75
& 34.60 & 63.88 & 76.10 & 17.73 \\

\rowcolor{gray!15}
ALBEF-4M + Ours
& 7.75\gain{+0.02} & 17.60\gain{+0.55} & 23.01\gain{+0.34} & 11.87\gain{+0.12}
& 34.65\gain{+0.05} & 65.45\gain{+1.57} & 79.18\gain{+3.08} & 18.98\gain{+1.25} \\

ALBEF-14M$_{(\mathit{finetuned})}$
& 8.60 & 18.91 & 25.60 & 13.16
& 35.11 & 63.86 & 77.37 & 18.93 \\

\rowcolor{gray!15}
ALBEF-14M + Ours
& 8.63\gain{+0.03} & 19.27\gain{+0.36} & 25.81\gain{+0.21} & 13.20\gain{+0.04}
& 35.12\gain{+0.01} & 67.18\gain{+3.32} & 80.16\gain{+2.79} & 20.73\gain{+1.80} \\

XVLM-4M$_{(\mathit{finetuned})}$
& 17.28 & 31.46 & 40.42 & 23.58
& 35.87 & 68.32 & 81.34 & 21.44 \\

\rowcolor{gray!15}
XVLM-4M + Ours
& 17.33\gain{+0.05} & \second{32.86}\gain{+1.40} & \second{41.51}\gain{+1.09} & 23.82\gain{+0.24}
& 35.98\gain{+0.11} & 68.47\gain{+0.15} & \second{81.75}\gain{+0.41} & 21.94\gain{+0.50} \\

XVLM-16M$_{(\mathit{finetuned})}$
& \second{17.85} & 32.18 & 41.15 & \second{24.13}
& \second{36.52} & \second{69.23} & 81.32 & \second{22.71} \\

\rowcolor{gray!15}
XVLM-16M + Ours
& \best{17.97}\gain{+0.12} & \best{34.15}\gain{+1.97} & \best{42.40}\gain{+1.25} & \best{24.59}\gain{+0.46}
& \best{36.74}\gain{+0.22} & \best{69.51}\gain{+0.28} & \best{82.59}\gain{+1.27} & \best{23.46}\gain{+0.75} \\

\hline

GeoText-1652~\cite{GeoText165206}
& \second{39.97} & \second{64.62} & \second{77.33} & \second{50.71}
& \second{39.25} & \second{70.24} & \second{82.48} & \second{24.00} \\

\rowcolor{gray!15}
GeoText-1652 + Ours
& \best{40.01}\gain{+0.04}
& \best{65.88}\gain{+1.26}
& \best{77.86}\gain{+0.53}
& \best{51.00}\gain{+0.29}
& \best{39.54}\gain{+0.29}
& \best{71.72}\gain{+1.48}
& \best{84.78}\gain{+2.30}
& \best{24.67}\gain{+0.67} \\

HCCM~\cite{ruan2025hccm33}
& 29.58 & 51.82 & 63.90 & 39.14
& 31.90 & 63.31 & 75.59 & 21.27 \\

\rowcolor{gray!15}
HCCM + Ours
& 29.59\gain{+0.01}
& 52.34\gain{+0.52}
& 64.71\gain{+0.81}
& 39.28\gain{+0.14}
& 31.90\same{+0.00}
& 64.07\gain{+0.76}
& 76.45\gain{+0.86}
& 21.44\gain{+0.17} \\

NGCG-MLLMs~\cite{chen2026turning}
& 19.41 & 39.65 & 51.46 & 28.17
& 17.98 & 44.55 & 59.48 & 9.46 \\

\rowcolor{gray!15}
NGCG-MLLMs + Ours
& 19.43\gain{+0.02}
& 40.18\gain{+0.53}
& 51.78\gain{+0.32}
& 28.24\gain{+0.07}
& 18.16\gain{+0.18}
& 49.82\gain{+5.27}
& 63.91\gain{+4.43}
& 9.85\gain{+0.39} \\

\bottomrule

\end{tabular}%

}

\label{tab100}

\vspace{-0.20in}

\end{table*}

\textit{Text Query.}
Under the text-query setting, natural language is the only query input. This setting is challenging because textual descriptions often provide incomplete spatial layouts, local entities, and viewpoint-dependent cues, causing the initial retrieval results to include many semantically relevant but geographically incorrect candidates. UniGeo mitigates this ambiguity through candidate-conditioned query refinement and candidate-level verification, which introduce discriminative evidence while preserving the original query intent.

As shown in Tables~\ref{tab1} and~\ref{tab100}, we evaluate UniGeo on both GeoText-1652 and UAVReason across three groups of methods, including directly evaluated pre-trained vision-language backbones, fine-tuned retrieval backbones, and recent task-specific geo-localization methods. For each baseline, UniGeo is applied in a plug-and-play manner without modifying its underlying retrieval architecture.
On GeoText-1652, under the pre-trained setting, UniGeo improves XVLM-16M from 5.27/10.71/14.29 to 5.50/13.27/18.40 in terms of R@1/R@5/R@10, while increasing mAP from 8.21 to 9.04. Under the fine-tuned setting, the improvement on XVLM-4M is more pronounced, increasing R@1/R@5/R@10 from 12.47/21.80/27.48 to 13.08/31.40/42.84, with gains of 9.60 and 15.36 percentage points at R@5 and R@10, respectively. Meanwhile, its mAP is improved from 17.95 to 21.10. For the task-specific GeoText-1652 baseline, UniGeo further improves R@1/R@5/R@10 from 13.97/25.20/31.78 to 14.72/33.80/45.37, and increases mAP from 20.35 to 23.18. Consistent improvements are also observed when UniGeo is integrated with recent methods such as HCCM~\cite{ruan2025hccm33} and NGCG-MLLMs~\cite{chen2026turning}.
Similar trends are also observed on UAVReason, indicating that the effectiveness of UniGeo is not limited to GeoText-1652. Compared with R@1, UniGeo achieves more consistent improvements in R@10, suggesting that its main advantage lies in refining the ranking of the candidate set and promoting more correct localization results into the Top-10. For example, under the pre-trained setting, UniGeo improves the R@10 of X-VLM-16M from 30.18 to 30.98; after fine-tuning, its R@10 is further improved from 41.15 to 42.40. Among the task-specific methods, the GeoText-1652 baseline achieves the strongest absolute performance on UAVReason, which may be related to the construction of this evaluation protocol. Temporally neighboring UAV frames are regarded as valid localization targets, while the generated textual descriptions emphasize scene-level semantics and spatial relations, which are relatively well aligned with the spatial-relation modeling adopted by GeoText-1652. On top of this strong baseline, UniGeo further improves R@10 from 77.33 to 77.86, while the R@10 of HCCM also increases from 63.90 to 64.71. Overall, the consistent Top-10 gains on UAVReason further demonstrate that UniGeo effectively improves the ranking and discrimination of highly similar candidates.

Overall, the consistent improvements on both benchmarks show that UniGeo is not tied to a specific retrieval backbone or dataset. In particular, the generally larger gains at R@5 and R@10 indicate that UniGeo mainly improves the ranking and discrimination of the leading candidate set, rather than merely altering the top-1 prediction. The consistent mAP improvements further suggest that the proposed post-retrieval reasoning process enhances the overall ordering of relevant candidates, providing more reliable candidate-level geo-localization.

\textit{Image Query.}
Under the image-query setting, a drone-view image is used as the query to retrieve the corresponding natural-language descriptions from the text gallery. Since the query modality differs from the primary text-query protocol, we use this setting as an auxiliary evaluation to examine the transferability of UniGeo across query modalities. In this setting, UniGeo does not perform text-query refinement, but instead applies lightweight candidate-level calibration to the initial candidates produced by the external retriever to improve image-text matching consistency.
As shown in Tables~\ref{tab1} and~\ref{tab100}, UniGeo consistently improves image-query performance on both datasets. On GeoText-1652, pre-trained XVLM-4M improves from 5.53/14.64/21.53 to 7.43/17.98/24.24 in terms of R@1/R@5/R@10, with mAP increasing from 10.00 to 12.34. XVLM-16M improves from 7.50/20.29/28.90 to 9.12/23.00/31.13, while mAP increases from 13.64 to 15.58. With the stronger GeoText-1652 baseline, UniGeo further improves R@1/R@5/R@10 from 28.81/56.45/69.30 to 29.32/56.80/69.57 and mAP from 40.60 to 41.98.
Similar trends are observed on UAVReason. For example, pre-trained XVLM-4M improves R@5/R@10 from 15.86/25.25 to 19.14/31.22, with mAP increasing from 1.83 to 2.31. XVLM-16M improves R@1/R@5/R@10 from 5.60/16.60/27.28 to 5.70/20.95/32.90, while mAP increases from 2.10 to 2.38. Under the fine-tuned setting, ALBEF-14M improves R@5/R@10 from 63.86/77.37 to 67.18/80.16, with mAP increasing from 18.93 to 20.73. For the task-specific GeoText-1652 baseline, UniGeo further improves R@1/R@5/R@10 from 39.25/70.24/82.48 to 39.54/71.72/84.78 and mAP from 24.00 to 24.67. Consistent improvements are also observed for HCCM and NGCG-MLLMs.
Overall, although image query is not the primary application protocol of UniGeo, it consistently improves candidate ranking on both datasets. In particular, the gains in R@5, R@10, and mAP on stronger baselines indicate that candidate-level calibration further improves the overall ranking of relevant texts within the candidate set.

\begin{table*}[t]
\centering
\small
\caption{\textbf{Evaluation of query refinement and grounded description quality.} We compare UniGeo with representative multimodal generation models on whole-image captioning and region-grounded description generation. Region-Match Acc@1 evaluates whether the generated description is correctly associated with the target region. Higher values indicate better performance for all metrics.}
\resizebox{\textwidth}{!}{%
\setlength{\tabcolsep}{5pt}
\renewcommand{\arraystretch}{1.15}
\begin{tabular}{lcccc|cccc}
\toprule
\multirow{2}{*}{Model} & \multicolumn{4}{c|}{Whole-Image Caption} & \multicolumn{4}{c}{Region-Grounded Description} \\
\cmidrule(lr){2-5} \cmidrule(lr){6-9}
& BLEU-1 $\uparrow$ & BLEU-2 $\uparrow$ & BLEU-4 $\uparrow$ & unigram F1 $\uparrow$
& Region-Match Acc@1 $\uparrow$ & BLEU-1 $\uparrow$ & BLEU-4 $\uparrow$ & F1 $\uparrow$   \\
\midrule
\rowcolor{gray!15}
Ours   & \textbf{0.42} & \textbf{0.30} & \textbf{0.16} & \textbf{0.48} & \textbf{0.71} & \textbf{0.27} & \textbf{0.06} & \textbf{0.34}  \\
UniLIP~\cite{tang2025unilip59} & 0.38 & 0.19 & 0.06 & 0.39 & 0.69 & 0.22 & 0.03 & 0.28 \\
Qwen-3.5-4B~\cite{qwen35omniblog}   & 0.29 & 0.14 & 0.03 & 0.34 & 0.70 & 0.19 & 0.02 & 0.26 \\
Bagel~\cite{deng2025bagel}  & 0.36 & 0.23 & 0.10 & 0.42 & 0.69 & 0.18 & 0.03 & 0.26 \\
\bottomrule
\end{tabular}}
\label{tab2}
\end{table*}

\begin{table*}[t]
\centering
\small
\renewcommand{\arraystretch}{1.15}
\setlength{\tabcolsep}{5pt}
\caption{\textbf{Evaluation of pose-aware cross-view image generation quality.} We compare UniGeo with representative multimodal generation models on two cross-view synthesis directions: drone generation and satellite generation. FID, PSNR, and SSIM measure generation quality, while MeanIdxErr additionally evaluates generation accuracy in the drone generation setting. Best and second-best results are highlighted in bold and underline, respectively.}
\resizebox{0.78\textwidth}{!}{%
\begin{tabular}{lcccc|ccc}
\toprule
\multirow{2}{*}{Model} 
& \multicolumn{4}{c|}{Drone Generation} 
& \multicolumn{3}{c}{Satellite Generation} \\
\cmidrule(lr){2-5}\cmidrule(lr){6-8}
& FID $\downarrow$ & PSNR $\uparrow$ & SSIM $\uparrow$ & MeanIdxErr $\downarrow$
& FID $\downarrow$ & PSNR $\uparrow$ & SSIM $\uparrow$
 \\
\midrule
\rowcolor{gray!15}
Ours 
& \textbf{34.45} & \textbf{15.74} & \textbf{0.18} & \textbf{15.40}
& \textbf{41.34} & \textbf{13.12} & \textbf{0.16}
\\
UniLIP~\cite{tang2025unilip59}
& \underline{50.72} & \underline{10.75} & 0.14 & 20.23
& \underline{56.72} & \underline{11.21} & \underline{0.13}
 \\
Qwen-Image-2511~\cite{wu2025qwenimagetechnicalreport}
& 128.04 & 10.29 & \underline{0.16} & \underline{19.62}
& 94.58 & 10.88 & \underline{0.13}
\\
Bagel~\cite{deng2025bagel}
& 208.18 & 7.31 & 0.05 & 21.79
& 113.55 & 7.82 & 0.05
 \\
\bottomrule
\end{tabular}%
}
\label{tab3}
\vspace{-0.20in}
\end{table*}

\textbf{Query Refinement and Grounded Description Quality.}
As shown in Table~\ref{tab2}, we evaluate the semantic generation quality of UniGeo from two perspectives: whole-image description and region-level grounded description.
The purpose of this experiment is to examine whether the unified vision-language backbone can generate discriminative semantic expressions relevant to geo-localization, rather than merely evaluating generic captioning capability.
For the whole-image description task, UniGeo achieves BLEU-1, BLEU-2, BLEU-4, and unigram F1 scores of 0.42, 0.30, 0.16, and 0.48, respectively, outperforming UniLIP~\cite{tang2025unilip59}, Qwen-3.5-4B~\cite{qwen35omniblog}, and Bagel~\cite{deng2025bagel} across all metrics. In particular, UniGeo improves BLEU-4 from 0.06 to 0.16 over UniLIP, indicating that the task-adapted UniGeo can not only cover more accurate semantic keywords, but also produce more stable phrase-level scene descriptions. For the region-level grounded description task, UniGeo also achieves the best performance, with a Region-Match Acc@1 of 0.71 and BLEU-1, BLEU-4, and F1 scores of 0.27, 0.06, and 0.34, respectively. These results demonstrate that UniGeo can more accurately associate generated descriptions with target regions and produce semantic expressions with stronger local specificity.
By generating more accurate whole-image semantic summaries and region-level grounded descriptions, UniGeo supplements the missing scene, object, and spatial cues in the original query, thereby providing more discriminative evidence for the subsequent candidate verification stage.

\textbf{Pose-Aware Cross-View Image Generation Quality.}
As shown in Table~\ref{tab3}, we assess the cross-view image generation quality of UniGeo in two directions: drone-view generation and satellite-view generation.
This experiment is designed to examine whether the unified vision-language backbone can learn geo-localization-oriented cross-view geometric transformations, rather than merely assessing generic image generation capability.
For the drone generation task, UniGeo achieves 34.45, 15.74, 0.18, and 15.40 in terms of FID, PSNR, SSIM, and MeanIdxErr, respectively, outperforming UniLIP~\cite{tang2025unilip59}, Qwen-Image-2511~\cite{wu2025qwenimagetechnicalreport}, and Bagel~\cite{deng2025bagel} across all metrics. In particular, compared with UniLIP, UniGeo substantially reduces FID from 50.72 to 34.45 and improves PSNR from 10.75 to 15.74, while also achieving better results in terms of SSIM and MeanIdxErr. These results indicate that the drone-view images generated by UniGeo not only exhibit higher visual fidelity, but also preserve target correspondence and structural consistency more accurately.
For the satellite generation task, UniGeo again delivers the best overall performance, achieving 41.34, 13.12, and 0.16 in FID, PSNR, and SSIM, respectively, all of which are clearly superior to those of the competing methods. Compared with UniLIP, UniGeo reduces FID from 56.72 to 41.34, while further improving both PSNR and SSIM. This suggests that UniGeo is likewise able to preserve the global semantics and spatial structure of the target region effectively in the reverse cross-view generation direction.

\begin{figure}[t]
    \centering
    
    \begin{subfigure}[t]{0.95\linewidth}
        \centering
        \includegraphics[width=\linewidth]{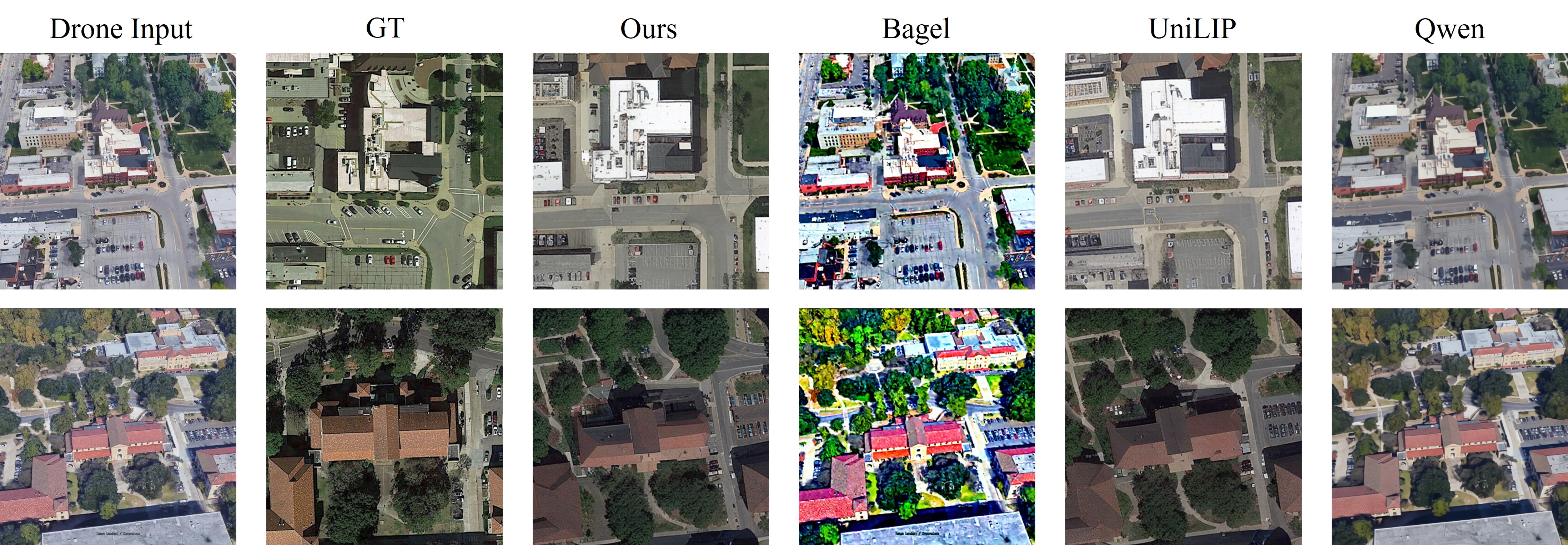}
        \caption{Drone-to-satellite generation.}
        \label{fig3a}
    \end{subfigure}
    
    \vspace{0.5em}
    
    \begin{subfigure}[t]{0.95\linewidth}
        \centering
        \includegraphics[width=\linewidth]{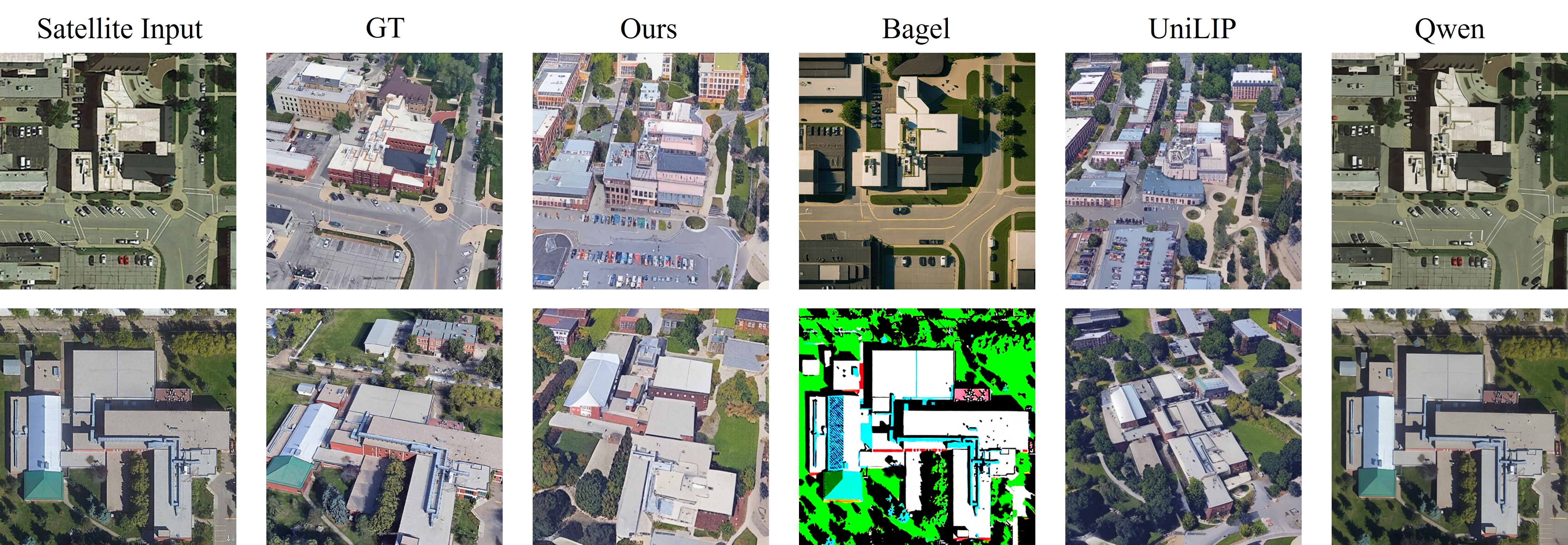}
        \caption{Satellite-to-drone generation.}
        \label{fig3b}
    \end{subfigure}

    \caption{\textbf{Qualitative comparison of bidirectional cross-view image generation.} 
    We compare UniGeo with other multimodal generation models on two synthesis directions. In satellite-view generation, the target image is generated under the default north-up orientation, while in drone-view generation, the viewing direction and pose are controlled by the KML metadata. }
    \label{fig3}
    \vspace{-0.20in}
\end{figure}

As shown in Fig.~\ref{fig3}, we observe that the qualitative results further support these findings. For drone-to-satellite generation, UniGeo better preserves road topology, building footprints, parking areas, and vegetation distribution under the default north-up satellite orientation. For satellite-to-drone generation, where the viewing direction and pose are controlled by KML metadata, UniGeo generates drone-view images with more coherent perspective and spatial layout. By contrast, Bagel often produces distorted or over-saturated textures, Qwen-Image tends to preserve the source-view appearance rather than performing a complete view transformation, and UniLIP shows weaker local structural consistency. These results indicate that UniGeo can generate cross-view samples with stronger semantic and geometric consistency, providing effective support for synthetic negative construction and candidate-level discriminative learning.

\begin{table*}[t]
\centering
\small
\renewcommand{\arraystretch}{1.12}
\caption{\textbf{Ablation study and hyperparameter sensitivity.} (a) evaluates the contribution of different components under the text-query and image-query settings, where CQR, CV, and HNL denote Candidate-Conditioned query refinement, candidate-level verification, and hard-negative learning, respectively. \textit{Refinement only} enables CQR alone and uses the refined text query for retrieval-side ranking without CV. \textit{Verifier only} enables CV alone on the baseline candidate pool without CQR. \textit{Hard Negative} applies CV with HNL, while \textit{w/o Hard Negative} removes HNL from the full framework.
(b) reports the sensitivity to the hyperparameters $K$ and $\alpha $ under the text-query setting. Best and second-best results are highlighted in bold and underline.}
\begin{minipage}[t]{0.65\textwidth}
\centering
\subcaption{Main mechanism ablation.}
\label{tab4a}
\resizebox{\linewidth}{!}{%
\begin{tabular}{lccc|ccc|ccc}
\toprule
\multirow{2}{*}{Method} 
& \multicolumn{3}{c|}{Components} 
& \multicolumn{3}{c|}{Text Query} 
& \multicolumn{3}{c}{Image Query} \\
\cmidrule(lr){2-4}\cmidrule(lr){5-7}\cmidrule(lr){8-10}
& CQR & CV & HNL & R@1 & R@5 & R@10 & R@1 & R@5 & R@10 \\
\midrule
Baseline            
& -- & -- & -- 
& 13.97 & 25.20 & 31.78 
& \underline{28.81} & \underline{56.45} & \underline{69.30} \\

+ Refinement only   
& \checkmark & -- & -- 
& 12.21 & 22.67 & 28.96 
& -- & -- & -- \\

+ Verifier only     
& -- & \checkmark & -- 
& 14.06 & 25.88 & 32.50 
& -- & -- & -- \\

+ Hard Negative     
& -- & \checkmark & \checkmark 
& 14.56 & 32.32 & 42.44 
& \textbf{29.32} & \textbf{56.80} & \textbf{69.57} \\

+ w/o Hard Negative   
& \checkmark & \checkmark & -- 
& \underline{14.67} & \underline{32.70} & \underline{42.70} 
& \underline{28.81} & \underline{56.45} & \underline{69.30} \\

\rowcolor{gray!12}
Ours (Full)               
& \checkmark & \checkmark & \checkmark 
& \textbf{14.72} & \textbf{33.80} & \textbf{45.37} 
& \textbf{29.32} & \textbf{56.80} & \textbf{69.57} \\
\bottomrule
\end{tabular}%
}
\end{minipage}
\hfill
\begin{minipage}[t]{0.285\textwidth}
\centering
\subcaption{Hyperparameter sensitivity.}
\label{tab4b}
\resizebox{\linewidth}{!}{%
\begin{tabular}{lccc}
\toprule
\multirow{2}{*}{Hyperparameter} & \multicolumn{3}{c}{Text Query} \\
\cmidrule(lr){2-4}
& R@1 & R@5 & R@10 \\
\midrule
\rowcolor{gray!12}
$K=16$       & \textbf{14.72} & \textbf{33.80} & \textbf{45.37} \\
$K=32$       & 14.68 & 33.73 & 45.35 \\
$K=64$       & 14.70 & 33.78 & 45.36 \\
\midrule
\rowcolor{gray!12}
$\alpha =0.1$  & \textbf{14.72} & \textbf{33.80} & \textbf{45.37} \\
$\alpha =0.05$ & 14.66 & 33.78 & 45.34 \\
$\alpha =0.2$  & 14.71 & 33.79 & 45.37 \\
\bottomrule
\end{tabular}%
}
\end{minipage}

\label{tab4}
\end{table*}

\begin{table*}[t]
\centering
\small
\renewcommand{\arraystretch}{1.12}
\setlength{\tabcolsep}{4pt}
\caption{\textbf{Analysis of hard-negative verification learning.}
(a) evaluates different negative sources for training the verification head. 
\textit{None} denotes the full framework without hard-negative verification learning, while \textit{Real HN} and \textit{Syn HN} denote real retrieval hard negatives and synthetic hard negatives, respectively. 
Avg. PairAcc denotes the average pairwise discrimination accuracy of the Stage-III verification head on held-out real and synthetic hard-negative pairs, and Gap denotes the absolute difference between the two pair accuracies. A smaller Gap indicates more balanced discrimination across different hard-negative sources. 
(b) studies the influence of the mixing ratio between real and synthetic hard negatives.}

\begin{minipage}[t]{0.55\textwidth}
\centering
\subcaption{Effect of hard-negative sources.}
\label{tab5a}
\resizebox{\linewidth}{!}{%
\begin{tabular}{lccccc}
\toprule
\multirow{2}{*}{Training Negatives} 
& \multicolumn{3}{c}{Text Query} 
& \multirow{2}{*}{Avg. PairAcc $\uparrow$}
& \multirow{2}{*}{Gap $\downarrow$} \\
\cmidrule(lr){2-4}
& R@1 & R@5 & R@10 & & \\
\midrule
None              
& 14.67 & 32.70 & 42.70 
& -- & -- \\

Real HN           
& \textbf{14.72} & \textbf{33.80}  & 45.32 
& 59.45 & 18.65 \\

Syn HN            
& \textbf{14.72} & 33.79  & 45.33 
& 69.50 & 44.07 \\

\rowcolor{gray!12}
Real + Syn HN     
& \textbf{14.72} & \textbf{33.80}  & \textbf{45.37} 
& \textbf{74.95} & \textbf{9.33} \\
\bottomrule
\end{tabular}%
}
\end{minipage}
\hfill
\begin{minipage}[t]{0.39\textwidth}
\centering
\subcaption{Effect of real/synthetic mixing ratio.}
\label{tab5b}
\resizebox{\linewidth}{!}{%
\begin{tabular}{lccc}
\toprule
\multirow{2}{*}{Real : Syn HN} 
& \multicolumn{3}{c}{Stage-III Head Evaluation} \\
\cmidrule(lr){2-4}
& Real Acc.$\uparrow$ & Syn Acc.$\uparrow$ & Gap$\downarrow$ \\
\midrule
$4:0$       & 72.09 & 93.64 & 21.55 \\
$3:1$       & 71.78 & 93.55 & 21.77 \\
$1:3$       & 70.06 & \textbf{97.58} & 27.52 \\
\rowcolor{gray!12}
$2:2$       & \textbf{73.29} & 94.55 & \textbf{21.26} \\
\bottomrule
\end{tabular}%
}
\end{minipage}

\label{tab5}
\vspace{-.10in}
\end{table*}

\subsection{Ablation Studies and Further Discussion}

\textbf{Effect of Candidate-Conditioned Disambiguation.}
As shown in Table~\ref{tab4a}, we analyze the effects of different key components of UniGeo on the geo-localization task. Here, Baseline denotes the performance of the original GeoText-1652~\cite{GeoText165206} retrieval backbone.
For text-query geo-localization, using query refinement alone reduces the performance from 13.97/25.20/31.78 to 12.21/22.67/28.96 in terms of R@1/R@5/R@10. This result indicates that the refined query should not be used as an independent replacement for the original query. In the absence of candidate-level verification constraints, query refinement may introduce additional information that is not fully
consistent with the true target, thereby interfering with the retrieval ranking. By contrast, introducing candidate-level verification alone improves the performance from 13.97/25.20/31.78 to 14.06/25.88/32.50, suggesting that candidate-level verification can provide a certain degree of fine-grained discrimination within the original candidate pool. However, the gains of this variant remain limited, indicating that candidate re-ranking alone is still insufficient to fully address the semantic incompleteness of textual queries and the confusion caused by highly similar candidates.
When the verification head is further trained with hard negatives, text-query geo-localization exhibits clear improvements, especially at larger retrieval depths, with R@1/R@5/R@10 increasing to 14.56/32.32/42.44. This suggests that hard negatives expose the verification head to more highly confusable candidates, thereby strengthening its ability to discriminate fine-grained cross-view differences. Meanwhile, the variant without hard-negative learning achieves 14.67/32.70/42.70 under the text-query setting, showing that the combination of query refinement and candidate-level verification already yields stable gains over the baseline. Building upon this, the complete UniGeo model further achieves the best overall performance of 14.72/33.80/45.37, reaching the highest values on all three text-query metrics. Compared with the GeoText-1652 baseline, the full UniGeo brings improvements of +0.75, +8.60, and +13.59 on R@1, R@5, and R@10, respectively. These findings indicate that the performance gains of UniGeo do not arise from any single component in isolation, but rather from the complementary synergy among query refinement, candidate-level verification, and hard-negative verification learning.
For image-query geo-localization, since the text-query refinement mechanism is not applicable, we primarily investigate the effect of hard-negative verification learning on candidate disambiguation in the image-query scenario. The results show that UniGeo improves the image-query baseline from 28.81/56.45/69.30 to 29.32/56.80/69.57 in terms of R@1/R@5/R@10. Notably, the results of the hard-negative variant and the full model are identical under the image-query setting, because query refinement is not involved in this protocol. These results indicate that hard-negative verification learning can also enhance candidate-level discriminative capability for image-based queries, although the improvement is more moderate than that observed in the text-query setting.

\textbf{Hyperparameter Study.}
We further evaluate the sensitivity of UniGeo to the local shortlist size $K$ and the fusion coefficient $\alpha$.
As shown in Table~\ref{tab4b}, UniGeo achieves the best performance at $K=16$, with R@1/R@5/R@10 reaching 14.72/33.80/45.37.
Further increasing $K$ to 32 or 64 brings negligible changes, since most competitive candidates are already covered by a compact local shortlist.
Similarly, $\alpha=0.1$ yields the best result, while nearby values such as $\alpha=0.05$ and $\alpha=0.2$ remain comparable.
This indicates that UniGeo performs conservative candidate-level calibration rather than aggressively overwriting the original retrieval order.
We therefore set $K=16$ and $\alpha=0.1$ as the default configuration.

\begin{figure}[t]
    \centering
    \includegraphics[width=0.9\linewidth]{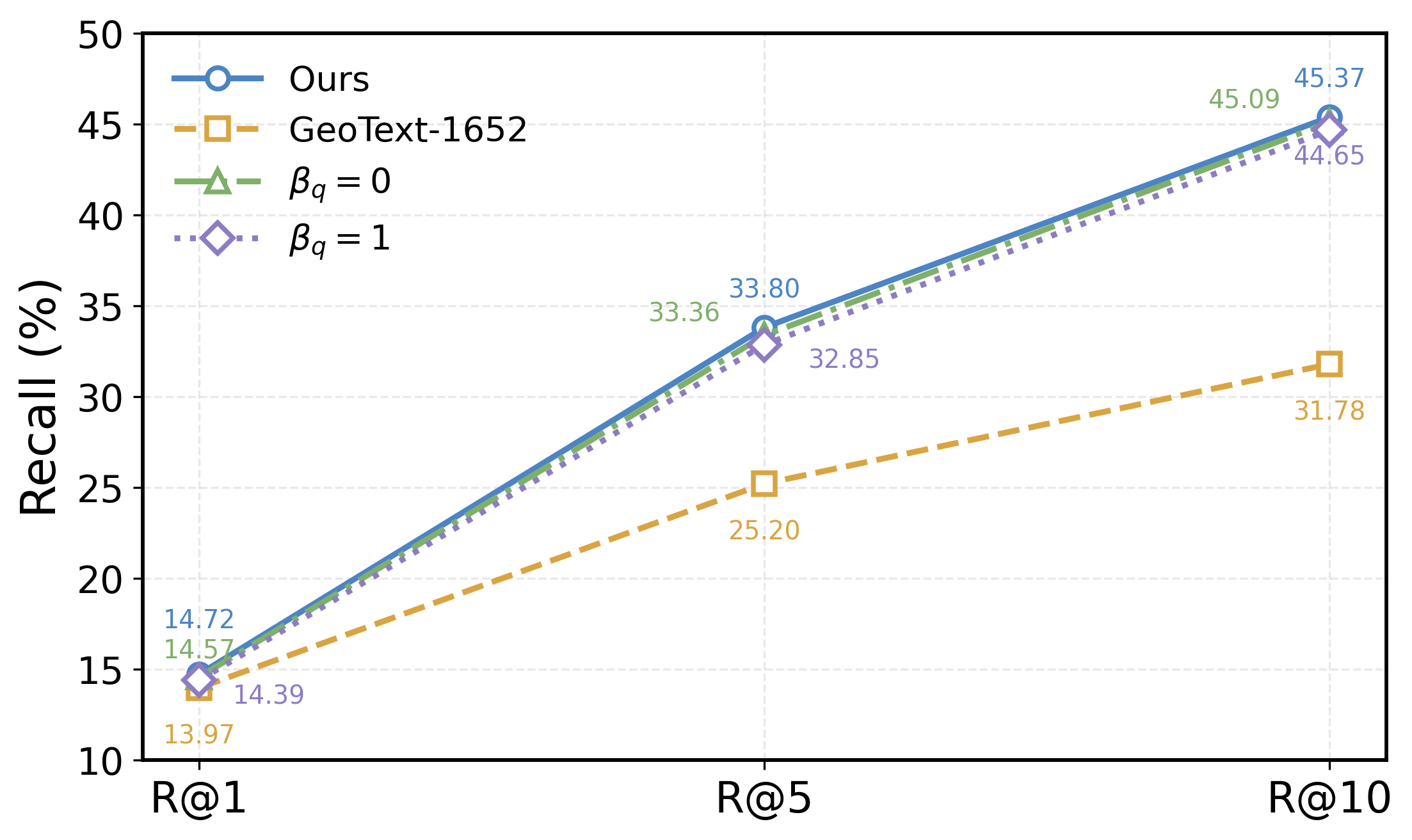}
    \caption{\textbf{Effect of different $\beta_q$ settings on retrieval performance.} We compare the baseline GeoText-1652 model, two variants with $\beta_q=0$ and $\beta_q=1$, and our model. Our model consistently achieves the best R@1, R@5, and R@10 performance.}
    \label{fig9}
    \vspace{-.20in}
\end{figure}

In addition, we analyze the effect of the branch-balancing coefficient $\beta_q$, as shown in Fig.~\ref{fig9}.
Here, $\beta_q=0$ denotes using only the original-query score $s_i^{(o)}$, while $\beta_q=1$ denotes using only the refined-query score $s_i^{(r)}$.
Compared with the GeoText-1652 baseline, both fixed settings achieve clear improvements, indicating that Candidate-conditioned query refinement and shared-pool scoring are effective under different query-branch configurations.
Specifically, $\beta_q=0$ achieves 14.57/33.36/45.09 in terms of R@1/R@5/R@10, while $\beta_q=1$ obtains 14.39/32.85/44.65.
The slightly better performance of $\beta_q=0$ suggests that the original-query branch still provides stable discriminative cues on this dataset.
Moreover, the full model achieves the best result of 14.72/33.80/45.37, showing that adaptively combining the two query branches leads to more reliable candidate scoring.
The small gap between the two fixed settings further indicates that UniGeo is relatively robust to the choice of $\beta_q$.

\textbf{Effect of Hard-Negative Sources.}
As shown in Table~\ref{tab5a}, we first examine the influence of different hard-negative sources for training the verification head. Compared with the setting without hard-negative learning, both Real HN and Syn HN improve R@5 and R@10, indicating that highly confusable negatives are beneficial for candidate-level verification. Real HN captures naturally retrieved near-miss samples, while Syn HN provides generation-induced confusable patterns. Combining the two sources achieves the best R@10 and the highest Avg. PairAcc, while also reducing the Gap between real and synthetic hard-negative discrimination. These results show that real and synthetic hard negatives provide complementary supervision for local candidate disambiguation.

We further evaluate the influence of the real/synthetic mixing ratio, as reported in Table~\ref{tab5b}. The results show that the \(2{:}2\) mixture achieves the highest Real Acc. and the lowest Gap, while maintaining competitive Syn Acc. This indicates that a balanced mixture helps the verification head avoid overfitting to either real retrieval errors or synthetic artifacts. Therefore, we adopt the Real+Syn strategy with a \(2{:}2\) mixing ratio in UniGeo.

\textbf{Effect of MLLM backbones.}
As shown in Table~\ref{tab8}, we evaluate the effect of different refinement models under the text-query setting by fixing the baseline retrieval model as GeoText-1652 and keeping the same Plug-and-Play Inference pipeline. When the refinement model is replaced with off-the-shelf InternVL3-2B, the performance improves from 13.97/25.20/31.78 to 14.69/32.01/42.51 in terms of R@1/R@5/R@10, with mAP increasing from 20.35 to 22.64. Replacing the refinement model with Qwen3-VL-2B further improves the results to 14.71/32.85/43.94 and 22.95 mAP. These results indicate that generic MLLMs can provide useful semantic refinement cues when integrated into the proposed plug-and-play inference pipeline. Nevertheless, the proposed UniGeo refinement model achieves the best performance of 14.72/33.80/45.37 and 23.18 mAP, consistently outperforming the off-the-shelf MLLM variants. The larger gains on R@5 and R@10 suggest that our refinement model provides more effective geo-semantic cues for promoting correct candidates into the leading candidate set, demonstrating that the performance improvement is not merely due to using a generic MLLM within the inference pipeline, but further benefits from the proposed task-oriented refinement design.

\begin{table}[t]
\centering
\small
\setlength{\tabcolsep}{5pt}
\caption{\textbf{ Ablation Study on the refinement model under the text-query setting.} The baseline retrieval model is fixed as GeoText-1652. ``InternVL3'' and ``Qwen3-VL'' denote replacing our refinement model with the off-the-shelf InternVL3-2B and Qwen3-VL-2B in the same Plug-and-Play Inference pipeline, while ``Ours'' denotes the proposed UniGeo refinement model.}
\begin{tabular}{llcccc}
\toprule
 Refinement Model & R@1 & R@5 & R@10 & mAP \\
\midrule

 -- (Baseline) 
& 13.97 & 25.20 & 31.78 & 20.35 \\

 InternVL3 (frozen)~\cite{zhu2025internvl3} 
& 14.69 & 32.01 & 42.51 & 22.64 \\

 Qwen3-VL (frozen)~\cite{bai2025qwen3} 
& \underline{14.71} & \underline{32.85} & \underline{43.94} & \underline{22.95} \\

\rowcolor{gray!12}
 Ours 
& \textbf{14.72} & \textbf{33.80} & \textbf{45.37} & \textbf{23.18} \\
\bottomrule
\end{tabular}
\label{tab8}
\end{table}


\begin{table}[t]
\centering
\small
\setlength{\tabcolsep}{6pt}
\caption{\textbf{Ablation study of the proposed pose-aware generation objective for drone-view image generation.} ``only diff'' denotes using only the diffusion reconstruction loss,
``w/o pose'' and ``w/o heading'' remove the pose consistency loss and heading auxiliary loss, 
``$\lambda_p=0.1$'' and ``$\lambda_p=0.3$'' only change the pose loss weight to 0.1 and 0.3, 
``$\lambda_h=0.2$'' only changes the heading auxiliary loss weight to 0.2.}
\begin{tabular}{lcccc}
\toprule
Method & FID $\downarrow$ & PSNR $\uparrow$ & SSIM $\uparrow$ & MeanIdxErr $\downarrow$ \\
\midrule
only diff      & \textbf{33.76} & \textbf{15.86} & \textbf{0.185} & 22.84 \\
w/o pose       & \underline{34.08} & \underline{15.82} & \underline{0.183} & 22.17 \\
w/o heading    & 34.73 & 15.69 & 0.178 & 18.36 \\
\midrule
$\lambda_p=0.1$        & 34.96 & 15.71 & 0.177 & 17.12 \\
$\lambda_p=0.3$       & 36.58 & 15.49 & 0.168 & \underline{15.86} \\
$\lambda_h=0.2$     & 35.37 & 15.60 & 0.173 & 16.09 \\
\midrule
\rowcolor{gray!12}
\textbf{Ours}  & 34.45 & 15.74 & 0.180 & \textbf{15.40} \\
\bottomrule
\end{tabular}
\label{tab6}
\end{table}

\textbf{Ablation Study on Pose-Aware Generation Objectives.} 
We further conduct an ablation study on the pose-aware generation module to analyze the effects of different generation objectives and weight settings. As shown in Table~\ref{tab6}, the only diff variant uses only the diffusion reconstruction objective and achieves favorable FID, PSNR, and SSIM results. However, it suffers from a much higher MeanIdxErr, indicating that image reconstruction alone can preserve visual fidelity but fails to provide accurate viewpoint control. When pose consistency is removed, MeanIdxErr increases from 15.40 to 22.17, demonstrating that the pose constraint is crucial for controllable cross-view generation. Removing the heading auxiliary loss increases MeanIdxErr to 18.36, suggesting that heading supervision mainly contributes to fine-grained directional calibration.

Further comparisons under different weight settings show that a weaker pose constraint is insufficient to fully regulate viewpoint generation, while a stronger pose constraint can reduce pose error but degrades FID, PSNR, and SSIM, indicating that overly strong geometric supervision may impair image naturalness. Similarly, increasing the heading weight alone does not outperform the full model. Overall, the diffusion reconstruction objective mainly preserves image quality, pose consistency provides global viewpoint control, and the heading auxiliary loss improves directional consistency. The full model achieves the lowest MeanIdxErr while maintaining competitive fidelity metrics, demonstrating that the proposed loss combination provides a better trade-off between generation quality and pose controllability.


\begin{figure}[t]
    \centering
    \includegraphics[width=0.9\linewidth]{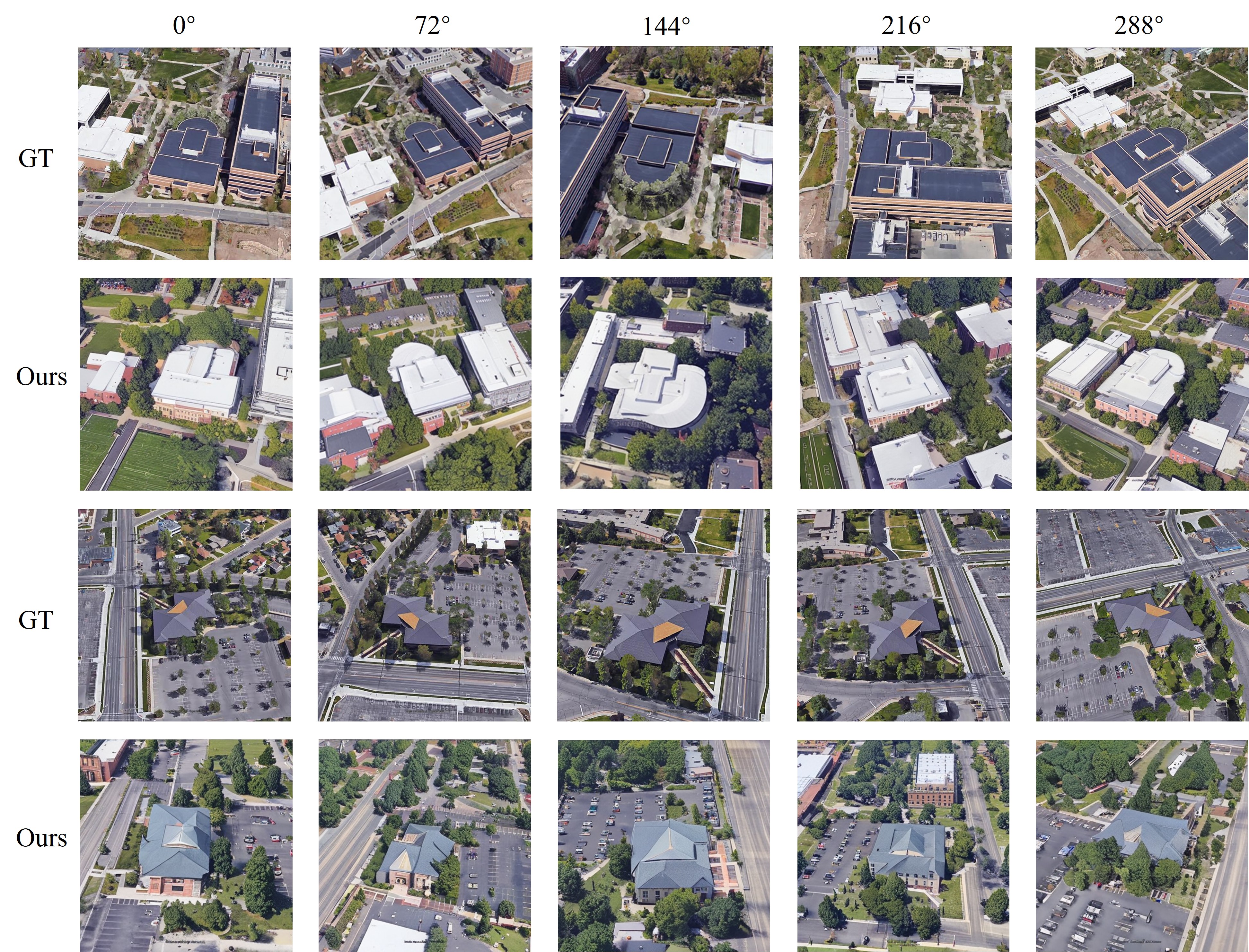}
    \caption{\textbf{Qualitative results of pose-aware cross-view generation.}
Rows 1 and 3 show real reference images, while Rows 2 and 4 show the corresponding generated results.
The viewing direction is controlled by the KML heading, and different columns correspond to different camera orientations.}
    \label{fig4}
    \vspace{-.20in}
\end{figure}

\textbf{Computational Cost Analysis.}
UniGeo is built on a frozen UniLIP~\cite{tang2025unilip59} backbone with approximately 4.18B parameters, which is shared across all candidates and can be reused without task-specific fine-tuning. It introduces only about 1.19M additional trainable parameters for localization, verification, and gating. UniGeo increases the average query time from 0.032 s to 0.415 s because it performs dynamic query refinement and candidate-level re-ranking over the top-$K_0$ candidates returned by the external retriever. In terms of computational cost, the XVLM retriever requires 106.42G FLOPs per query, while UniGeo requires approximately 2.6T FLOPs due to dynamic query refinement and candidate-level verification. Since query refinement is dynamically generated, the FLOPs of UniGeo are approximate, with 331.97G FLOPs corresponding to output-token decoding. This overhead is incurred once per query at the inference stage and does not affect the high-recall candidate retrieval process. In exchange for this overhead, UniGeo improves R@10 by 13.6 percentage points without modifying the retrieval backbone, making it suitable for accuracy-critical geo-localization scenarios where reliable candidate-level disambiguation is more important than the additional inference overhead.

\textbf{Qualitative Analysis of Pose-Aware Cross-View Generation.}
We visualize pose-aware cross-view generation results under different KML heading conditions in Fig.~\ref{fig4}.
As the heading varies from $0^\circ$ to $288^\circ$, the generated results show clear directional changes while preserving major geo-semantic elements, including buildings, roads, parking areas, and surrounding vegetation.
This indicates that the heading condition provides effective geometric control and enables the generation branch to capture plausible cross-view scene structures.

Although the generated images are not strictly identical to the real references in fine-grained details, such as roof shapes, local textures, and road boundaries, these discrepancies do not undermine their role in our framework.
The generation branch is used to construct geographically plausible and semantically confusable samples under controlled viewpoint changes, rather than to serve as a pixel-level reconstruction module.
These generated samples provide harder supervision than random negatives and help the verification head learn fine-grained distinctions among highly similar candidates.
Therefore, the qualitative results support the effectiveness of pose-aware cross-view generation for hard-negative verification learning.



\begin{figure*}[t]
\centering
\subfloat[Region-grounded captioning.]{
\includegraphics[height=1.44in]{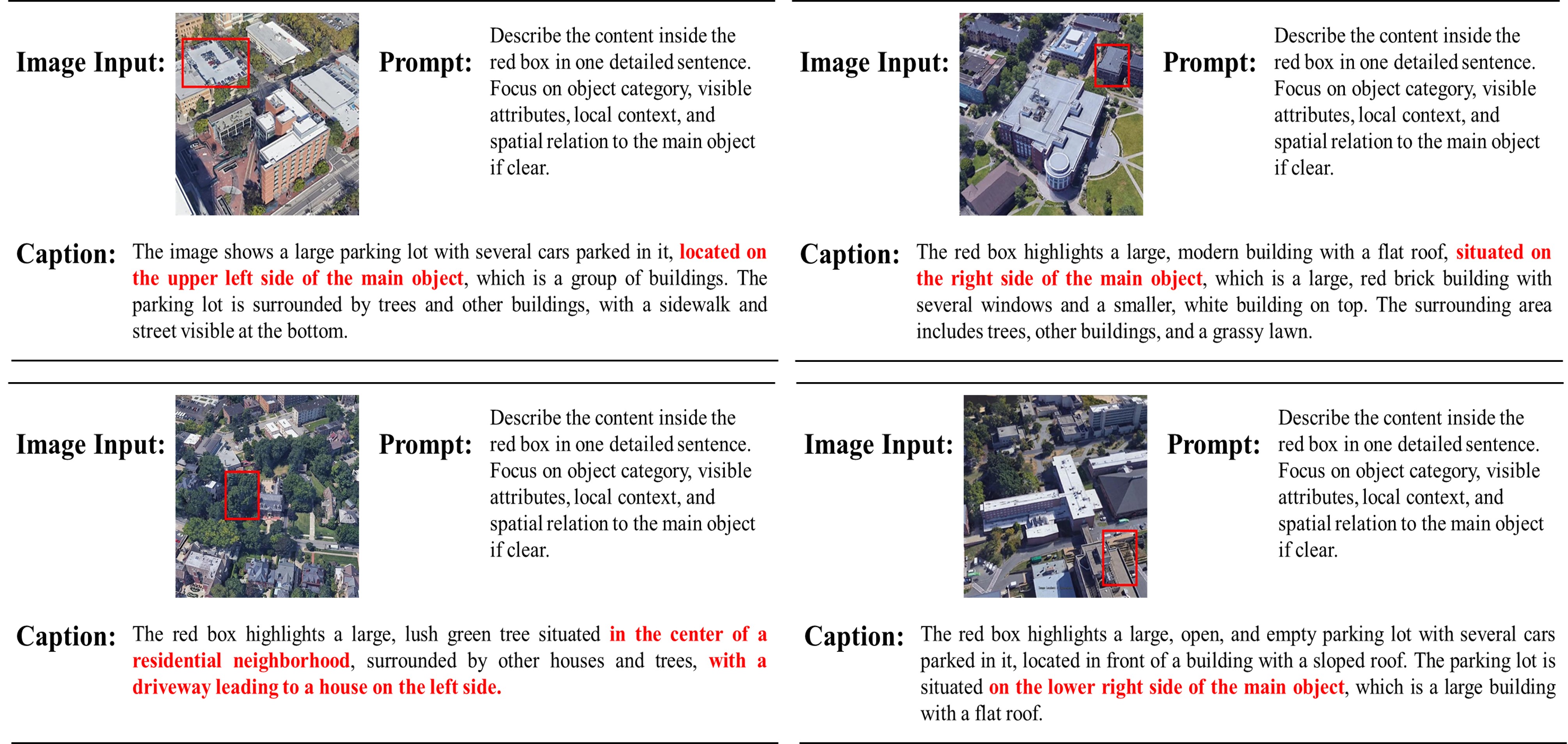}
\label{fig5}
}
\hfill
\subfloat[Text-query geo-localization.]{
\includegraphics[height=1.44in]{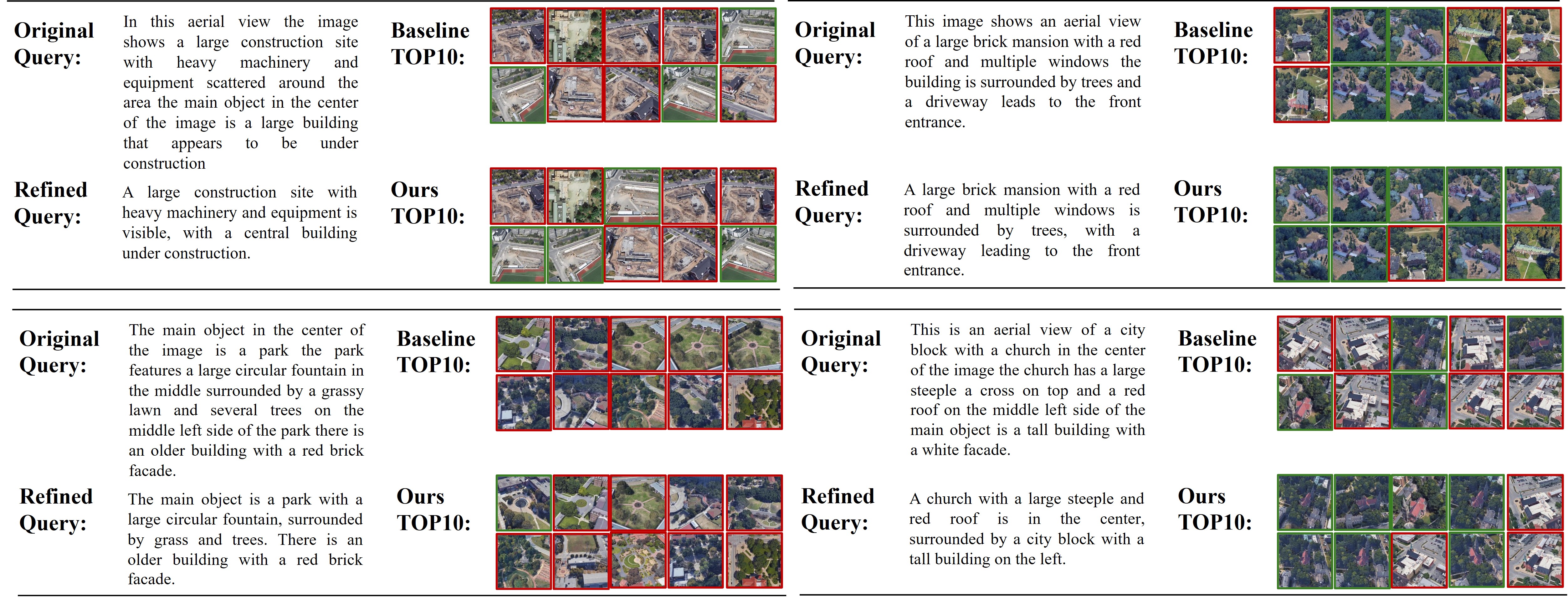}
\label{fig6}
}

\caption{\textbf{Qualitative results of geo-semantic understanding and text-query geo-localization.}
(a) Given an image and its corresponding bounding boxes, UniGeo generates region-level grounded descriptions.
(b) Each example presents the original query, the refined query, and the Top-10 retrieval results produced by the baseline and our method.
Green boxes indicate correct candidates, and red boxes indicate incorrect candidates. (Best
viewed when zooming in.)}
\label{fig:qualitative_results}
\vspace{-.10in}

\end{figure*}

\textbf{Qualitative Analysis of Geo-Semantic Understanding.}
The qualitative results in Fig.~\ref{fig5} illustrate the region-level geo-semantic understanding capability of UniGeo.
Given a red-boxed region in an aerial image, UniGeo generates localized descriptions that capture object categories, visual attributes, surrounding context, and spatial relations.
For example, it can identify geo-semantic elements such as parking lots, buildings, trees, and roads, and describe their relative positions within the broader aerial scene.
These results show that UniGeo does not merely recognize isolated local objects, but can associate region-level semantics with global scene context, thereby providing structured spatial-semantic cues for subsequent geo-localization.

\textbf{Qualitative Results of Text-Query Geo-Localization.}
As shown in Fig.~\ref{fig6}, we further compare the Top-10 retrieval results of the Baseline and UniGeo under text-query geo-localization.
Each example includes the original query, the refined query generated by UniGeo, and the retrieval results of both methods, where green boxes denote correct candidates and red boxes denote incorrect ones.
The original queries often contain redundant descriptions and unstable relative-position expressions, whereas the refined queries preserve the core geo-semantic information and emphasize localization-relevant cues, such as target category, scene layout, road structure, vegetation distribution, and building attributes.

Compared with the Baseline, which mainly retrieves candidates with coarse semantic similarity, UniGeo improves either the ranking positions or the number of correct candidates within the Top-10 results.
This suggests that query refinement and candidate-level verification can provide more discriminative evidence for distinguishing highly similar geographic candidates.
Some incorrect candidates remain when different locations share similar buildings, roads, or vegetation patterns, indicating that text-query geo-localization is still challenging under severe candidate ambiguity.
Overall, the qualitative results show that UniGeo benefits from structured geo-semantic understanding and improves the reliability of text-query geo-localization.

\subsection{Discussion}
\label{sec:discussion}
The experimental results show that UniGeo brings much larger gains on R@5 and R@10 than on R@1 under the text-query setting. This pattern reflects the nature of text-guided geo-localization, where textual queries in GeoText-1652 usually provide region-level cues rather than instance-specific evidence that uniquely identifies the target. As multiple candidate regions may share similar road structures, building appearances, and spatial layouts, aggressively altering the top-1 ranking can be unstable, especially when the refined query contains uncertain or over-specified semantics. UniGeo therefore adopts a conservative candidate-level strategy: it preserves the original retrieval backbone as the candidate generator and refines the relative ranking among plausible candidates through query refinement, candidate verification, and hard-negative learning. The substantial improvements on R@5 and R@10 indicate that UniGeo effectively promotes correct targets that are originally ranked lower into the leading candidate set, thereby reducing the risk of missing correct locations in subsequent map inspection, human verification, or fine-grained localization. These observations suggest that UniGeo mainly improves shortlist-level recall coverage, while exact first-rank localization under sparse textual descriptions remains a challenging direction for future work.

\section{Conclusion}
\label{sec:conclusion}
This paper presents UniGeo, a unified geospatial vision-language framework for text-guided drone geo-localization. We argue that text-driven geo-localization is not merely a cross-modal retrieval problem, but also a post-retrieval geo-semantic reasoning problem. Natural-language queries often fail to fully describe the target region, while candidate geographic regions frequently exhibit similar building appearances, road structures, and spatial layouts. Motivated by this observation, UniGeo integrates geo-semantic understanding, pose-aware cross-view generation, and candidate-level verification into a unified pipeline, enabling the model to further mine fine-grained and discriminative spatial-semantic cues within the initial candidate pool. Experimental results show that UniGeo consistently improves the performance of multiple representative retrieval backbones on text-query geo-localization. Further analyses on region-level understanding, controllable generation, and hard-negative verification demonstrate its geospatial reasoning capability. Overall, UniGeo provides a valuable exploration toward unified geospatial vision-language intelligence for aerial scenarios.

\bibliography{reference}
\bibliographystyle{IEEEtran}

\begin{IEEEbiography}
[{\includegraphics[width=1in,height=1.25in,clip,keepaspectratio]{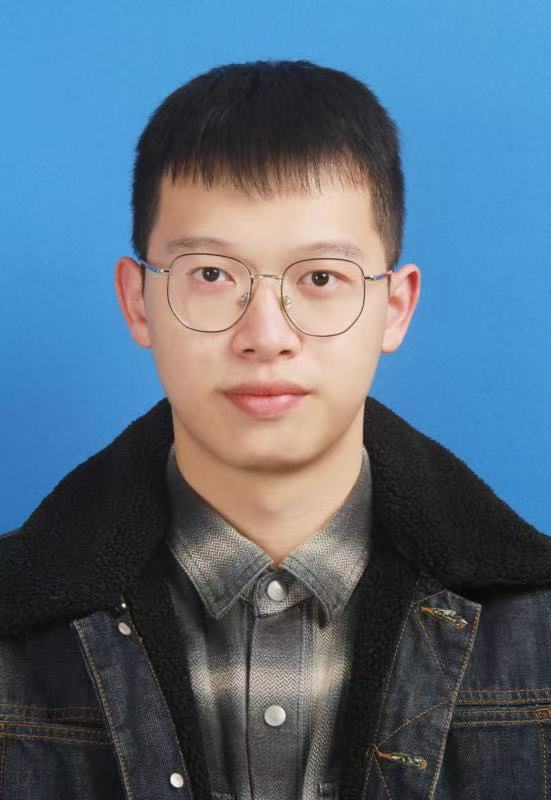}}]
{Jiahao Wen} is currently pursuing a Ph.D. degree in computer science with the School of Computer Engineering and Science, Shanghai University, Shanghai, China. His research interests include image retrieval, graph neural networks and meta-learning.
\end{IEEEbiography}

\vspace{-8mm}
\begin{IEEEbiography}
[{\includegraphics[width=1in,height=1.25in,clip,keepaspectratio]{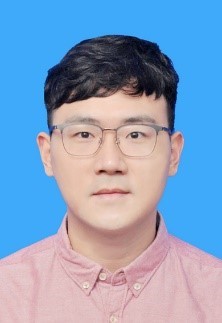}}]
{Hang Yu} is a Professor at the School of Computer Engineering and Science, Shanghai University,
China. He received his Ph.D. degree from the University of Technology Sydney, Australia, in 2020. He
was awarded the Outstanding Academic Leader of
Shanghai. His research interests include streaming
data mining, concept drift, and fuzzy systems. He
has authored or co-authored more than 60 publications and his publications have appeared in the IEEE
Transactions on Knowledge and Data Engineering,
IEEE Transactions on Neural Networks and Learning Systems, IEEE Transactions on Cybernetics and IEEE Transactions on Fuzzy Systems. He also regularly serves as a program committee member for numerous national and international conferences.
\end{IEEEbiography}

\vspace{-8mm}
\begin{IEEEbiography}
[{\includegraphics[width=1in,height=1.25in,clip,keepaspectratio]{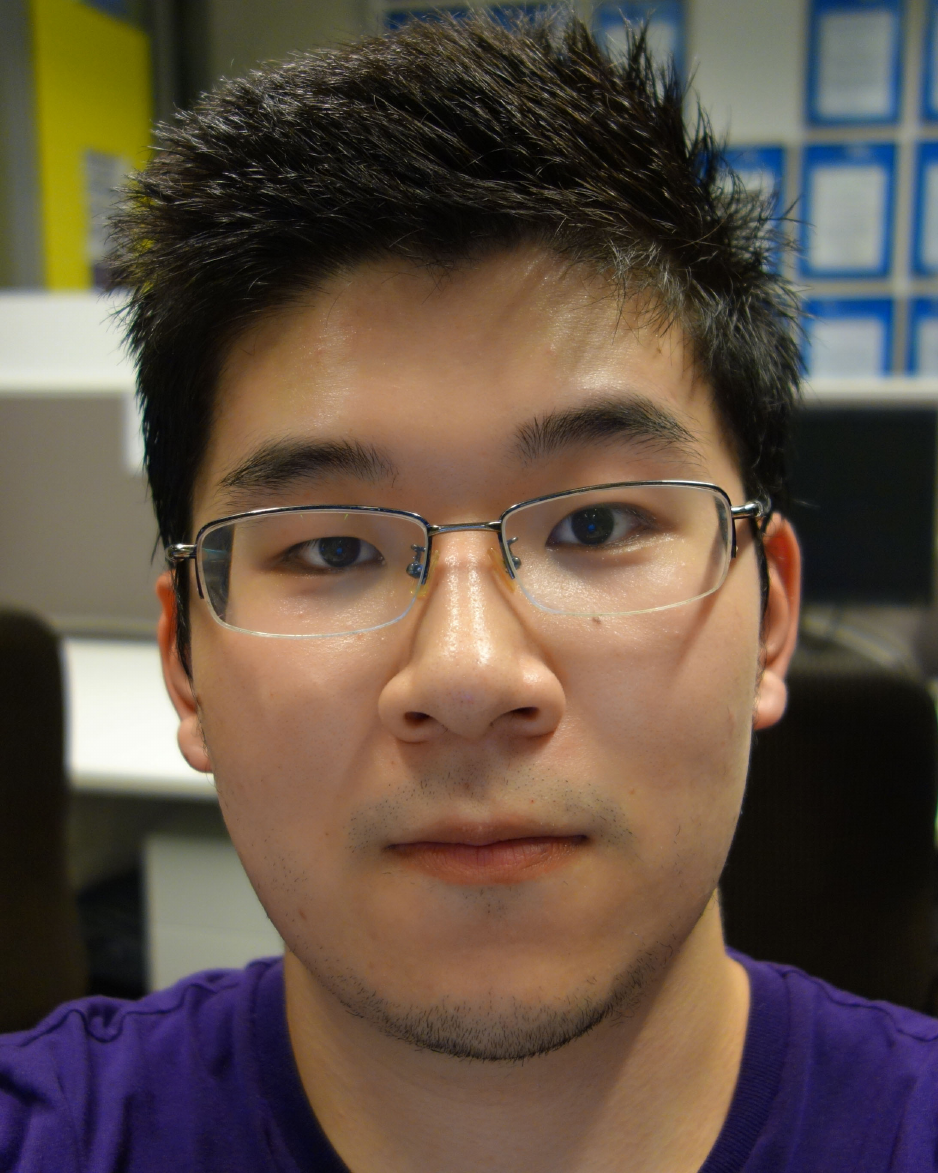}}]
{Zhedong Zheng} is an Assistant Professor with the University of Macau. He received the Ph.D. degree from the University of Technology Sydney in 2021 and the B.S. degree from Fudan University in 2016. He was a postdoctoral research fellow at the School of Computing, National University of Singapore. He received the IEEE Circuits and Systems Society Outstanding Young Author Award of 2021. His research interests include AIGC, Data-centric AI, and Spatial Intelligence. He actively serves the academic community, acting as a Senior PC for IJCAI and AAAI, an Area Chair for ACM MM'24, ACM MM'25 and ICASSP'25, and the Publication Chair for ACM MM'25 and AVSS'25.

\end{IEEEbiography}

\vfill

\end{document}